\documentclass{article} %
\usepackage{iclr2023_conference,times}

\usepackage{amsmath,amsfonts,bm}

\def\eqref#1{equation~\ref{#1}}

\def\1{\bm{1}}

\def\rmW{{\mathbf{W}}}

\def\vr{{\bm{r}}}
\def\vs{{\bm{s}}}

\DeclareMathAlphabet{\mathsfit}{\encodingdefault}{\sfdefault}{m}{sl}
\SetMathAlphabet{\mathsfit}{bold}{\encodingdefault}{\sfdefault}{bx}{n}

\DeclareMathOperator*{\argmin}{arg\,min}

\usepackage[utf8]{inputenc} %
\usepackage[T1]{fontenc}    %
\usepackage[colorlinks,
    linkcolor=red,citecolor=citecolor,urlcolor=blue]{hyperref}       %
\usepackage{url}            %
\usepackage{booktabs}       %
\usepackage{amsfonts}       %
\usepackage{nicefrac}       %
\usepackage{microtype}      %
\usepackage{graphicx}
\usepackage{colortbl}
\usepackage{multirow}
\usepackage{amsmath}
\usepackage{bm}
\usepackage{bbm}
\usepackage{amsthm}
\usepackage{harpoon}
\usepackage[]{algpseudocode}
\usepackage{enumitem} %
\usepackage[subrefformat=parens,labelformat=parens]{subfig}

\usepackage{wrapfig}
\usepackage[algo2e, ruled, vlined, linesnumbered, noend]{algorithm2e}
\usepackage{media9}
\usepackage{animate}

\definecolor{citecolor}{RGB}{34,139,34}
\definecolor{mydarkblue}{rgb}{0,0.08,1}
\definecolor{mydarkgreen}{rgb}{0.02,0.6,0.02}
\definecolor{mydarkred}{rgb}{0.8,0.02,0.02}
\definecolor{mydarkorange}{rgb}{0.40,0.2,0.02}
\definecolor{mypurple}{RGB}{111,0,255}
\definecolor{myred}{rgb}{1.0,0.0,0.0}
\definecolor{mygold}{rgb}{0.75,0.6,0.12}
\definecolor{myblue}{rgb}{0,0.2,0.8}
\definecolor{mydarkgray}{rgb}{0.,0.2,0.2}

\definecolor{lightred}{RGB}{255,235,235}
\definecolor{lightgreen}{RGB}{235,255,235}
\definecolor{lightblue}{RGB}{235,235,255}
\definecolor{lightcyan}{RGB}{235,255,255}
\definecolor{lightmagenta}{RGB}{255,235,255}
\definecolor{lightyellow}{RGB}{255,255,235}

\definecolor{qxkcolor}{RGB}{215,235,255}
\definecolor{softmaxcolor}{RGB}{230,235,255}
\definecolor{probxvcolor}{RGB}{255,255,235}

\definecolor{topkcolor}{RGB}{255,235,235}
\definecolor{zecolor}{RGB}{255,255,235}
\definecolor{dynacolor}{RGB}{235,255,255}

\definecolor{reviewcolor}{RGB}{0,0,200}

\newcommand{\calO}{\mathcal{O}}
\newcommand{\calF}{\mathcal{F}}

\newcommand{\calL}{\mathcal{L}}
\newcommand{\calP}{\mathcal{P}}
\newcommand{\calS}{\mathcal{S}}
\newcommand{\calT}{\mathcal{T}}
\newcommand{\calR}{\mathcal{R}}

\theoremstyle{plain}

\theoremstyle{definition}

\algdef{SE}[DOWHILE]{Do}{doWhile}{\algorithmicdo}[1]{\algorithmicwhile\ #1}%

\graphicspath{{./figs/}}

\newcommand{\name}{\texttt{HEAT}\xspace}

\title{
 HEAT: \underline{H}ardware-\underline{E}fficient \underline{A}utomatic \underline{T}ensor Decomposition for Transformer Compression
}

\author{Jiaqi Gu$^2$, Ben Keller$^1$, Jean Kossaifi$^1$, Anima Anandkumar$^{1,3}$, Brucek Khailany$^1$, David Z. Pan$^2$ \\
$^1$NVIDIA, $^2$The Univesity of Texas at Austin, $^3$Caltech\\
\texttt{jqgu@utexas.edu} \\
}

\iclrfinalcopy

\begin{document}

\maketitle
\begin{abstract}
\label{abstract}
Transformers have attained superior performance in natural language processing and computer vision.
Their self-attention and feedforward layers are overparameterized, 
limiting inference speed and energy efficiency.
Tensor decomposition is a promising technique to reduce parameter redundancy by leveraging tensor algebraic properties to express the parameters in a factorized form. 
Prior efforts used manual or heuristic factorization settings without hardware-aware customization, resulting in poor hardware efficiencies and large performance degradation. 

In this work, we propose a hardware-aware tensor decomposition framework, dubbed \name, that enables efficient exploration of the exponential space of possible decompositions and automates the choice of tensorization shape and decomposition rank with hardware-aware co-optimization.
We jointly investigate tensor contraction path optimizations and a fused Einsum mapping strategy to bridge the gap between theoretical benefits and real hardware efficiency improvement.
Our two-stage knowledge distillation flow resolves the trainability bottleneck and thus significantly boosts the final accuracy of factorized Transformers. 
Overall, we experimentally show that our hardware-aware factorized BERT variants reduce the energy-delay product by $5.7 \times$ with less than $1.1\%$ accuracy loss and achieve a better efficiency-accuracy Pareto frontier than hand-tuned and heuristic baselines.

\end{abstract}

\section{Introduction}
\label{sec:Introduction}

\begin{figure}[b]
    \centering
    \vspace{-15pt}
    \includegraphics[width=\columnwidth]{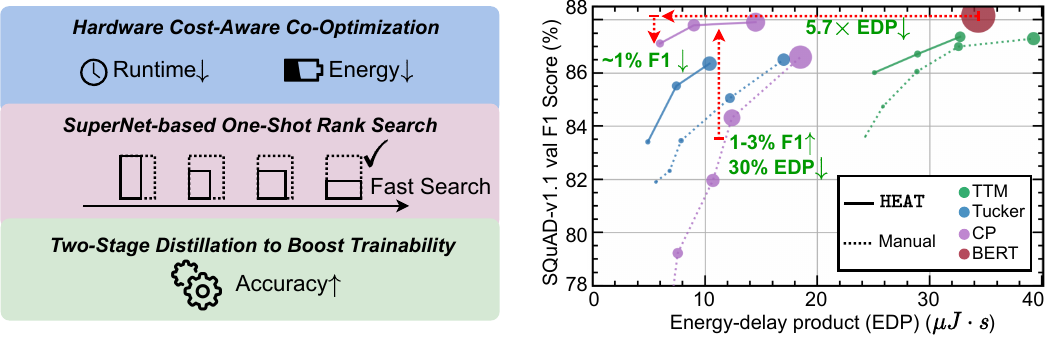}
    \vspace{-20pt}
    \caption{Hardware-efficient tensor decomposition framework \name achieves better accuracy-efficiency trade-off than manual settings.}
    \label{fig:Teaser}
\end{figure}

Transformer models have demonstrated record-breaking performance on natural language processing (NLP) tasks~\citep{NN_NAACL2018_Peters,NAACL2019_Devlin,NN_NeurIPS2020_Brown}.
However, the linear projection layers in multi-head self-attention (MHSA) and feedforward networks (FFNs) contain a large number of parameters that limit the efficient deployment of Transformers.
Therefore, compressing large-scale Transformers is an essential problem in practical NLP tasks.

\begin{figure*}
    \centering
    \subfloat[]{\includegraphics[width=0.31\columnwidth]{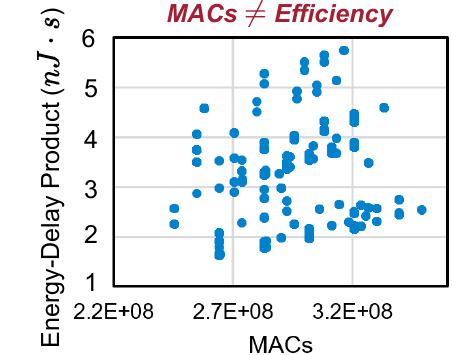}
    \label{fig:Motivation_MAC_EDP}
    }
    \subfloat[]{\includegraphics[width=0.31\columnwidth]{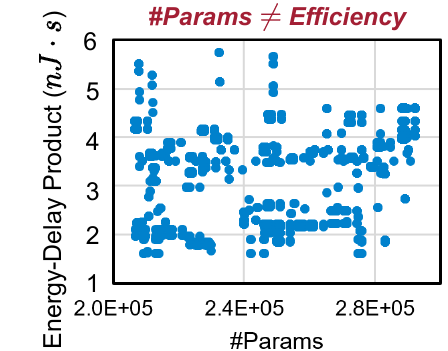}
    \label{fig:Motivation_Params_EDP}
    }
    \hspace{8pt}
    \subfloat[]{\includegraphics[width=0.276\columnwidth]{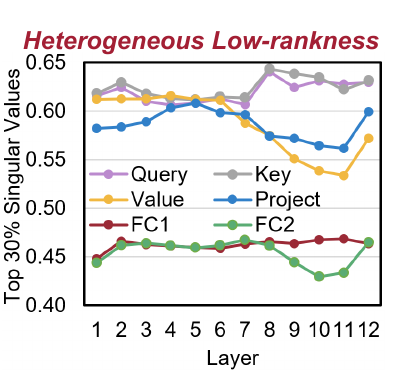}
    \label{fig:Motivation_Lowrank}
    }
    \vspace{-10pt}
    \caption{
    (a)(b) Plot different tensorization shapes and ranks in the space.
    Computations (MACs) and parameter counts are not accurate indicators of real hardware cost (\emph{lower energy-delay product means the application consumes less energy and runs faster~\citep{EDP}}).
    (c) Different tensors in BERT-base show different low-rankness, thus requiring per-tensor customization.
    }
    \vspace{-10pt}
    \label{fig:Motivation}
\end{figure*}

Tensor decomposition~\citep{kolda2009tensor} leverages the higher-order structure in tensors to efficiently express it in a factorized form, e.g., CP~\citep{CP}, tucker~\citep{Tucker}, tensor-train (TT)~\citep{NN_SIAM2011_Oseledets} decomposition.
It has found many applications in computer vision~\citep{panagakis2021tensor}. 
Tensor factorization can be applied to matrices by first \emph{tensorizing}~\citep{anandkumar2014tensor,NN_NIPS2015_Novikov} them (reshaping them into a higher-order tensor) and then factorizing that tensor. 
Among the many model compression methods~\citep{NN_AAAI2022_Zhen,NN_EMNLP2020_Hrinchunk,NN_NeruIPS2018_Zhen,NN_CVPR2022_Zhen}, it is a particularly promising approach to exploring the intrinsic redundancy of NN weight matrices. 
Prior work has successfully applied this method to reduce the number of parameters and computations in NNs by manually selecting the tensorization shapes and ranks~\citep{NN_NIPS2015_Novikov,NN_TSP2017_Sidiropoulos,NN_EMNLP2020_Hrinchunk,NN_NeruIPS2018_Zhen,NN_JMLR2020_Kossaifi, NN_ISCA2019_Deng,NN_CVPR2021_Yin,NN_IJCNN2018_Tjandra, NN_ICCV2021_Gu}.

However, three critical issues remain unresolved.
First, \textbf{improvements in compression metrics do not necessarily translate to better hardware efficiency}, a distinction ignored by most prior work.
As shown in Fig.~\ref{fig:Motivation_MAC_EDP} and \ref{fig:Motivation_Params_EDP}, number of multiply-accumulate operations (MACs) and parameters are imprecise indicators of energy efficiency and execution speed on real hardware.
This implies that a high compression ratio may not translate to real hardware efficiency benefits.
Moreover, we observe \textbf{heterogeneous low-rank characteristics} in different weight matrices in Fig.~\ref{fig:Motivation_Lowrank}, but previous methods ignore this heterogeneity and manually assign a global setting to all matrices based on heuristics~\citep{NN_ISCA2019_Deng,NN_EMNLP2020_Hrinchunk,NN_NeruIPS2018_Zhen, tensorly}, failing to explore the huge design space.
An additional challenge of factorized Transformers is the non-trivial performance drop after tensor decomposition.
Direct re-training cannot recover the accuracy of factorized Transformers due to the \textbf{optimization difficulty} of cascaded tensor contractions, which hinders their practical deployment.

To solve these challenges, we propose \name, a hardware-efficient tensor decomposition framework that features automated tensor decomposition with hardware-aware optimization, shown in Fig.~\ref{fig:Teaser}.
\name can efficiently explore the huge design space of tensorization while significantly improving the hardware efficiency based on the following approaches:
(1) compared to prior hardware-unaware tensor decomposition work, \name incorporates hardware cost feedback in the tensorization optimization flow to find expressive and hardware-efficient tensorization settings;
(2) instead of manually selecting a global rank setting via trial-and-error, \name leverages the heterogeneous low-rank property of different tensors and adopts a novel Rank SuperNet-based method to automatically search for efficient per-tensor rank settings in the exponentially large space with one-shot re-training cost;
(3) \name resolves the trainability challenge of factorized Transformers by introducing a two-stage knowledge distillation flow to significantly boost the task performance.

Based on the approaches in \name, we make the following contributions:
\begin{itemize}[leftmargin=*]
\setlength{\itemindent}{0.5em}
\item We deeply investigate the hardware efficiency of tensor decomposition-based model compression methods and propose an automatic framework for hardware-efficient Transformer factorization.
\item We move beyond conventional compression metrics and incorporate hardware cost into optimization to find hardware-efficient tensorization settings.
\item We propose a novel Rank SuperNet to explore the exponential space of heterogeneous per-tensor rank settings to push forward the accuracy-efficiency Pareto front with one-shot re-training cost.
\item We discuss the trainability bottleneck of factorized Transformer and resolve it via a two-stage distillation recipe to remedy the task performance degradation from factorization.
\end{itemize}

Our searched factorized BERT models outperform the original BERT with an estimated 5.7$\times$ lower energy-delay product (EDP) and surpass hand-tuned and heuristic baselines with 25\%-30\% lower EDP and 1-3\% accuracy improvement on SQuAD-v1.1 and SST-2 datasets.

\section{\name Automatic Tensor Decomposition Framework}
\label{sec:Method}

\subsection{Understanding Hardware-Efficient Tensor Decomposition}
\label{sec:Formulation}
\begin{wrapfigure}[20]{rH}{0.33\columnwidth}
\begin{minipage}{0.33\textwidth}
    \centering
    \vspace{-15pt}
    \includegraphics[width=\columnwidth]{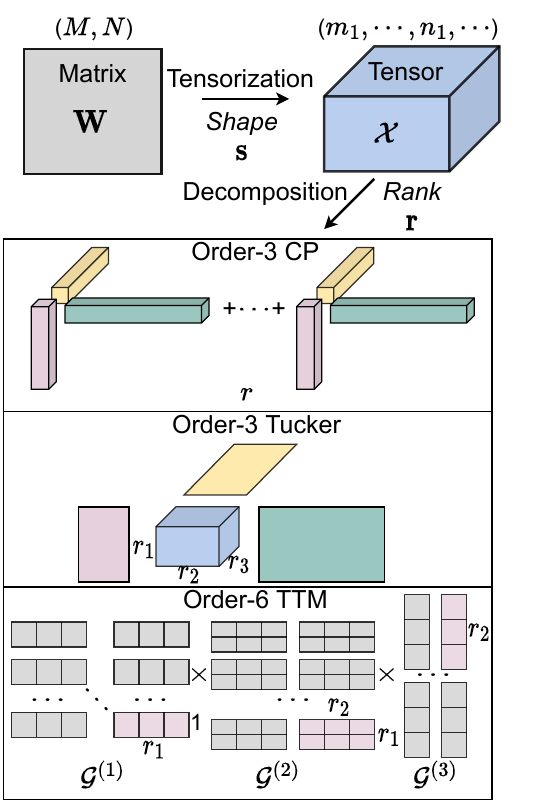}
    \vspace{-20pt}
    \caption{Illustration of tensor decomposition.}
    \label{fig:Decomposition}
\end{minipage}
\end{wrapfigure}

\noindent\textbf{Tensor Decomposition}.~
We first briefly introduce the basics of tensor decomposition.
As shown in Fig.~\ref{fig:Decomposition}, a matrix $\rmW\in\mathbb{R}^{M\times N}$ is tensorized into a high-order tensor $\mathcal{X}$ and further approximated by the product or summation of a series of smaller core tensors.
Representative decompositions include CP~\citep{CP}, Tucker~\citep{Tucker}, and tensor-train matrix (TTM)~\citep{NN_SIAM2011_Oseledets}.
For example, the order-$d$ TTM decomposition is formulated by

\begin{equation}
\small
\begin{aligned}
    \mathcal{X}((i_1,&j_1),\cdots\!,(i_d,\!j_d))\!=\!\mathcal{G}^{(1)}\!((i_1,j_1),:)\cdots\mathcal{G}^{(d)}(:,(i_d,j_d)),
\end{aligned}
\end{equation}
where each $\mathcal{G}^{(i)}\in\mathbb{R}^{r_{i-1}\times m_i\times n_i \times r_i}$ is called a core tensor, the size of tensorized $\mathcal{X}$ is called tensorization shape, i.e., $\vs=(m_1,\cdots,n_1,\cdots)$, where $M=\prod_{i} m_i$ and $N=\prod_{i} n_i$.
The variable dimensions of cores are called decomposition ranks, i.e, $\vr=(r_0,r_1,\cdots,r_{d-1},r_d)$.
The compression ratio is $c=\sum_i r_{i-1}m_in_ir_i/MN$.

To find a hardware-efficient decomposition, the goal is to determine the tensorization shape $\vs$ and rank $\vr$ for each matrix $\rmW$ that minimizes energy-delay product while maintaining high accuracy.

\noindent\textbf{Design Space}.~
The design space of possible ($\vs$, $\vr$) pairs is exponentially large.
We use Tucker decomposition as an example.
Given an $M\times N$ matrix $\rmW$, we assume its candidate factorization orders are $D=(d_1,\cdots,d_i,\cdots,d_k)$.
For any order $d_i\in D$, we define $\calS_i$ as all possible tensorization shapes, and we have $|\calS_i|=\calO((d_i!)^2)$.
For each tensorization shape $\vs_j=(m_1,\cdots,m_{d_i/2},n_1,\cdots,n_{d_i/2})\in\calS_i$, we denote the set of total possible ranks as $\calR_{ij}$, and one example rank setting is $\vr=(r_1,\cdots,r_{d_i})\in\calR_{ij}$.
There are $\calO(\sum_{i,j}|\calR_{ij}|)\approx\calO(\sum_i MN(d_i!)^2)$ different factorization settings for this matrix.
If we wish to explore per-tensor factorization settings for a DNN with $L$ weight matrices, then the complexity explodes to $\calO(\prod_l^L(\sum_i M^lN^l(d_i!)^2))$.
For example, the total possible decompositions for BERT-base are roughly $10^{632}$.
Exploring this huge combinatorial design space via brute-force search is intractable, especially considering that it requires costly model re-training and hardware cost simulation to evaluate each factorization shape-rank pair ($\vs$, $\vr$).

\noindent\textbf{Formulation}.~
To make this intractable problem efficiently solvable, we formulate the search as a \emph{three-level hierarchical optimization} as follows,
\begin{equation}
\small
    \begin{aligned}
        &\texttt{Level 3}:~\text{Train factorized model:}~ \Theta^*(\vs^*,\vr^*)=\argmin_{\Theta}\calL(\Theta(\vs^*,\vr^*),\mathcal{D}_{trn}),\\
        &~~~\texttt{Level 2}:~\text{Search rank:}~ \vr^*=\argmin_{\vr}~\big(1-\texttt{Acc}(\Theta^*(\vs^*,\vr))\big)\texttt{Cost}^*(\vs^*,\vr|\alpha)^{\gamma},\\
        &\quad\quad\quad\quad\quad\quad\quad\quad\quad\quad\Theta^*(\vs^*,\vr)=\argmin_{\Theta}\calL(\Theta(\vs^*,\vr),\mathcal{D}_{trn}),\\
        &~~~~~~\texttt{Level 1}:~\text{Search shape:}~ \vs^*=\texttt{Pareto}_{\vs}\big(\texttt{Cost}^*(\vs,\vr|\alpha),\epsilon,c)\\
        &\epsilon=\|\rmW-\rmW'(\vs,\vr)\|_{\calF}/\|\rmW\|_{\calF},~\texttt{Cost}^*(\vs,\vr|\alpha)=\min_{m}~\min_{p}~\texttt{Cost}(\vs,\vr,m,p|\alpha).
    \end{aligned}
\end{equation}
In \texttt{Level~1}, given an accelerator architecture $\alpha$, we find a Pareto optimal tensorization shape $\vs^*$ that minimizes decomposition error $\epsilon$, compression ratio $c$, and the minimum hardware cost $\texttt{Cost}^*$ obtained by optimizing tensor contraction path ($p$) and hardware mapping ($m$).
We use a standard \emph{energy-delay product (EDP)} as the hardware cost to reflect both energy consumption and runtime cost.
Then in \texttt{Level~2}, we search for optimal per-tensor rank settings $\vr^*$ while minimizing hardware cost and maximizing the task-specific performance.
In \texttt{Level~3}, we train the factorized Transformer model with the optimal ($\vs^*$, $\vr^*$) settings to find its optimal parameters $\Theta^*(\vs^*,\vr^*)$.

\subsection{The Proposed \name Framework}
\label{sec:ProposedFramework}
\begin{figure}[h]
    \centering
    \includegraphics[width=\columnwidth]{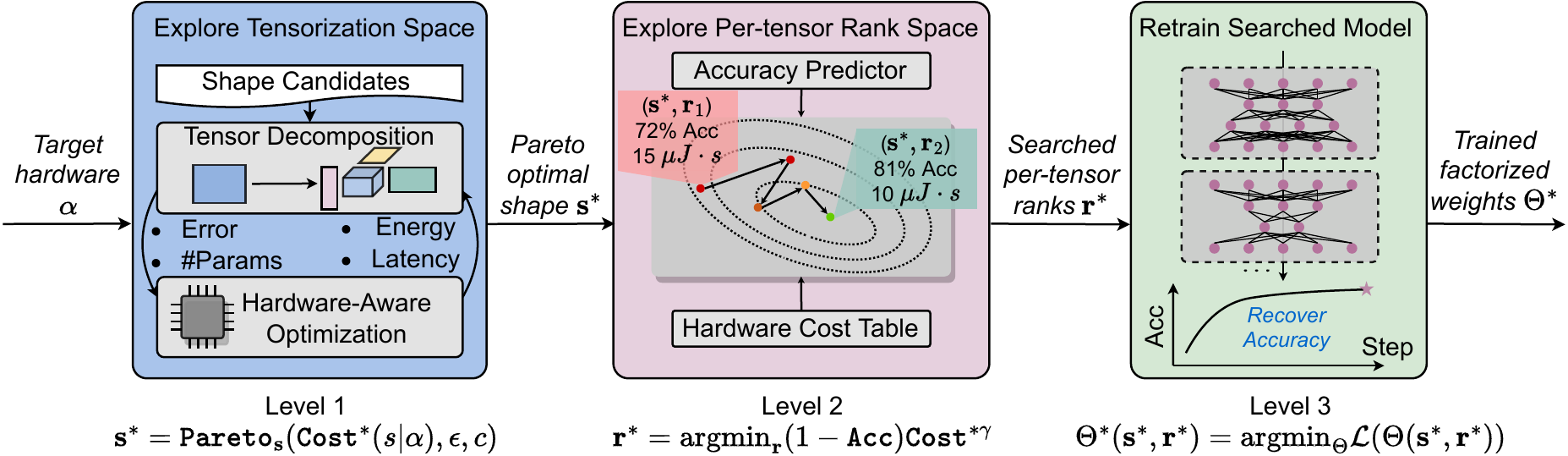}
    \caption{Overview of our hardware-efficient automatic tensor decomposition framework \name.
    }
    \label{fig:Overview}
\end{figure}

To solve this three-level optimization problem efficiently, we propose a three-stage framework \name, summarized in Fig.~\ref{fig:Overview}.
In the first stage, we simulate each shape candidate on a given inference accelerator architecture $\alpha$ and build a \textit{hardware cost table} for all shapes $\calT:\calS\times\calR\rightarrow\texttt{Cost}$, from which we select one Pareto optimal shape $\vs^*$ with low decomposition error $\epsilon$, low compression ratio $c$,
and low hardware cost. 
All matrices with the same size share this tensorization shape.
We import the optimal shape $\vs^*$ to the second stage, a one-shot rank search flow that efficiently explores per-tensor rank settings with minimum model re-training cost. 
With the searched optimal shape and rank ($\vs^*$, $\vr^*$), we enter the last step, a knowledge distillation-based re-training flow to recover the accuracy of factorized Transformer.

\subsubsection{\texttt{Level} 1: Pareto Optimal Tensorization Shape Search}
\label{sec:ShapeSearch}
\begin{wrapfigure}[15]{rH}{0.68\columnwidth}
\begin{minipage}{0.68\textwidth}
    \centering
    \vspace{-15pt}
    \includegraphics[width=\columnwidth]{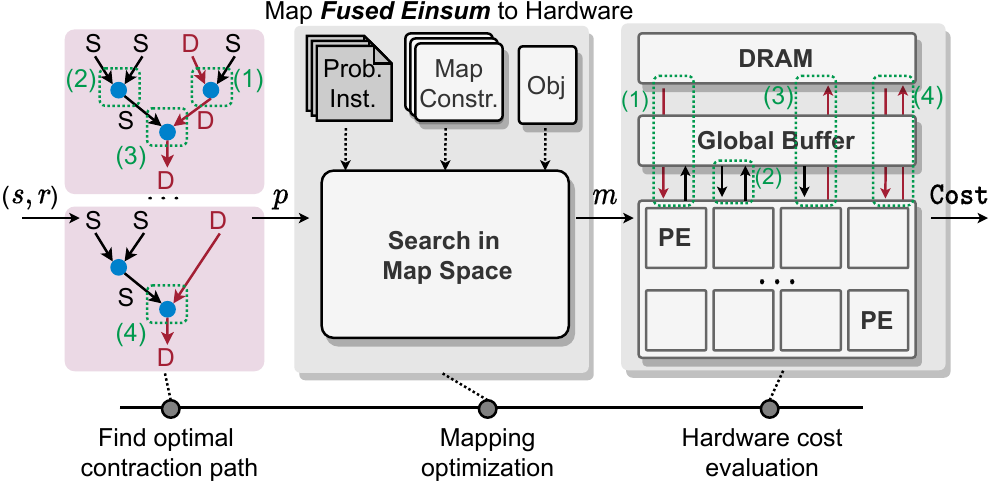}
    \vspace{-10pt}
    \caption{Efficiently map fused Einsum to \texttt{Simba-L} architecture aware of four memory access types.}
    \label{fig:FusedEinsum}
\end{minipage}
\end{wrapfigure}

The shape of the high-order tensor $\mathcal{X}$ is critical to the approximation error and hardware efficiency.
Unlike prior work that empirically selects the tensorization shape based on heuristics, we wish to find one that can achieve low approximation error with minimal hardware cost, which corresponds to multi-objective optimization.
To determine the real hardware cost of a given tensorization, as shown in Fig.~\ref{fig:FusedEinsum}, we must find a near-optimal contraction path $p$ and map it to the hardware $\alpha$ with an optimal mapping $m$.

\begin{figure}[htp]
    \centering
    \vspace{-10pt}
    \includegraphics[width=0.8\columnwidth]{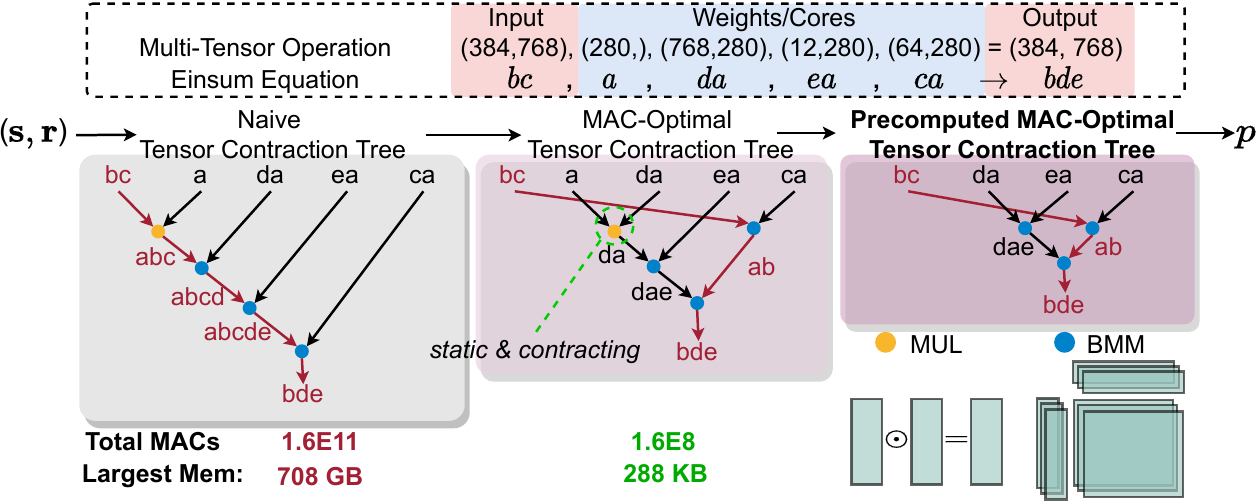}
    \vspace{-10pt}
    \caption{Our pre-computed MAC-optimal contraction path can significantly save hardware cost.}
    \label{fig:TensorContractionOptimize}
    \vspace{-10pt}
\end{figure}

\noindent\textbf{Tensor Contraction Path Optimization}.~
A factorized linear layer requires a series of tensor contractions, which can be described by a symbolic Einsum equation, as shown in Fig.~\ref{fig:TensorContractionOptimize}.
The order in which these tensors are contracted, or the \emph{contraction path, is critical to hardware efficiency}.
For example, we show a CP factorized layer in Fig.~\ref{fig:TensorContractionOptimize}.
The (768$\times$768) matrix is first reshaped to an order-3 (768$\times$12$\times$64) tensor.
Given a rank of 280, this tensor is decomposed into a length-280 weight vector and three CP cores.
Multiplying the input $x$ with CP cores corresponds to this Einsum equation: \texttt{bc,a,da,ea,ca}$\rightarrow$\texttt{bde}.
A naive tensor contraction path of this equation is shown on the left tree of Fig.~\ref{fig:TensorContractionOptimize}.
Each node represents a 2-operand tensor operation.
Simply following the left-to-right association order leads to considerable computation and intermediate storage overhead due to the many outer product operations.
In contrast, a MAC-optimal path with a more efficient association order reduces hardware cost by orders of magnitude~\citep{opt_einsum}.
Note that not all nodes in the tree need to be calculated on the fly. 
We can pre-compute \emph{static} and \emph{contracting} nodes to eliminate redundant memory and computation cost, e.g., element-wise multiplication and batched inner products.
Static nodes mean their inputs are known before inference, and contracting nodes mean those operations reduce the tensor size.
Hence, we only need to implement the pre-computed MAC-optimal path on the hardware accelerator.

\noindent\textbf{Map Fused Einsum to Hardware}.~
To efficiently implement factorized tensor operations, we customize our accelerator \texttt{Simba-L}~\citep{NN_MICRO2019_Shao} to perform a 1024-element matrix-vector multiplication each cycle to achieve high utilization on factorized Einsum workloads. 
However, 
\begin{wrapfigure}[29]{rH}{0.51\columnwidth}
\begin{minipage}{0.51\textwidth}
    \centering
    \vspace{-7pt}
    \includegraphics[width=\columnwidth]{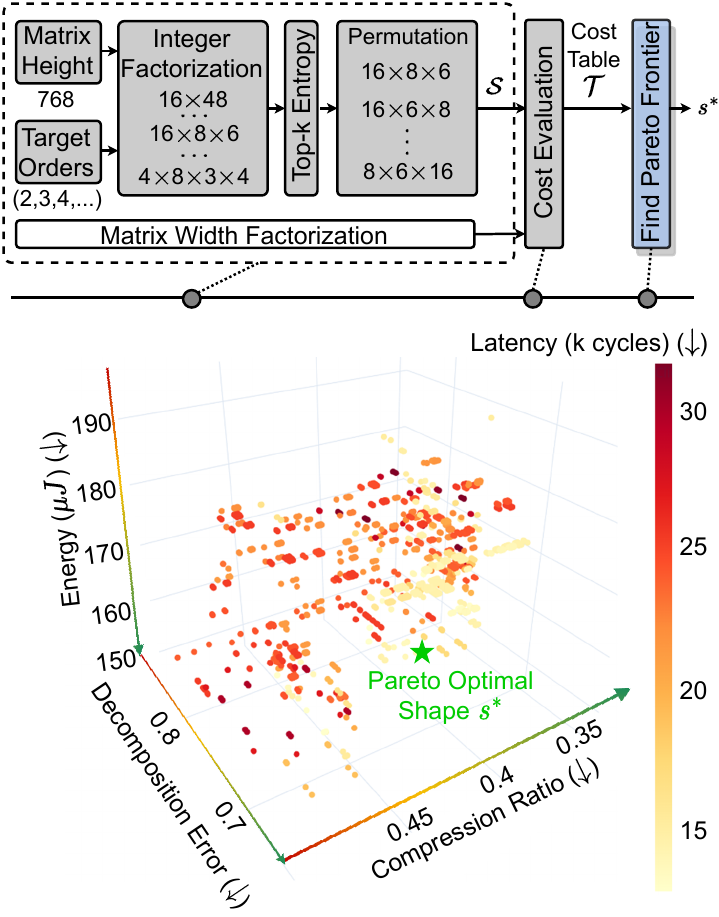}
    \vspace{-15pt}
    \caption{\textit{Top}: Generate shape candidates and find the Pareto optimal shape $\vs^*$ based on the cost table $\calT$.
    \textit{Bottom}: The \emph{Energy-Latency-Error-Ratio} space with $\sim$3000 candidate shape-rank pairs.}
    \label{fig:ShapeSpace}
\end{minipage}
\end{wrapfigure}
individually implementing each 2-operand node in $p$ would realize minimal efficiency benefit since the intermediate tensors are always stored in DRAM, introducing nontrivial data movement cost.
Instead, we implement a fused Einsum, minimizing redundant DRAM accesses by storing intermediate results in on-chip SRAM whenever possible to obtain the minimum hardware cost $\texttt{Cost}^*(\vs,\vr|\alpha)$ (See Appendix~\ref{sec:AppendixFusedEinsum}).
We use Timeloop~\citep{NN_ISPASS2019_Parashar} to search for an efficient mapping $m$ while minimizing energy-delay product, map the fused Einsum to the accelerator, and evaluate the energy and runtime.

\noindent\textbf{Search for the Pareto Optimal Shape: $\vs^*$}.~
We construct a search space for shape candidates.
In Fig.~\ref{fig:ShapeSpace}, we perform integer factorization on the matrix height/width, select top-k shapes with maximum entropy since we prefer uniform shapes across axes, and permute the axes to form the shape candidates $\calS$.
Then we decompose the matrix with all shape candidates to get the approximation error and evaluate their hardware costs to form a cost table $\calT$ (See details in Appendix~\ref{sec:AppendixCostSimulation}).
We visualize the cost table in Fig.~\ref{fig:ShapeSpace}.
We automatically detect the points on the Pareto-optimal surface and heuristically select the best shape $\vs^*$ with the lowest decomposition error.
We repeat this process for all different sizes of matrices in the model, 
completing the \texttt{Level 1} optimization.

\subsubsection{\texttt{Level} 2: Weight-Sharing Per-Tensor Rank Optimization}
\label{sec:RankSearch}
\begin{figure}
    \centering
    \includegraphics[width=0.85\columnwidth]{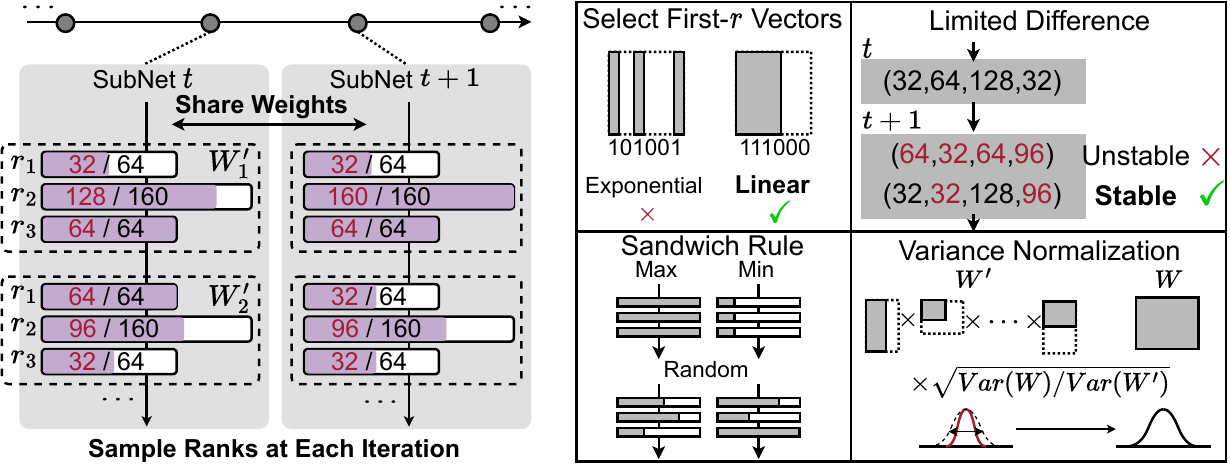}
    \vspace{-5pt}
    \caption{\textit{Left}: Rank SuperNet training flow.
    \textit{Right}: Illustration of training stabilization techniques.}
    \label{fig:SuperNetTrain}
    \vspace{-10pt}
\end{figure}

\noindent\textbf{One-Shot Rank Search via Weight-Sharing SuperNet}.~
The challenges in the \texttt{Level 2} optimization are twofold: (1) the exponentially large per-tensor rank search space and (2) the prohibitive cost of accuracy evaluation on a shape-rank pair.
These barriers make it impossible to select the best rank by enumeration.
Inspired by the high efficiency in weight-sharing neural architecture search (NAS)~\citep{NN_CVPR2019_Wu,NN_ICLR2020_Cai}, we propose a \textbf{Rank SuperNet} where each SubNet corresponds to a per-tensor rank setting.
As shown in Fig.~\ref{fig:SuperNetTrain}, at each iteration, we randomly sample valid ranks for each tensor, and different SubNets share the same set of parameters.
Hence, we can efficiently explore a large space of different rank settings.
Training this SuperNet can easily suffer from instability issues due to a large rank sampling variance, so we adopt four techniques to stabilize the SuperNet convergence.
(1) We only sample the first $r$ vectors along each axis to reduce the search space.
(2) We limit the rank difference across iterations to reduce variance.
(3) We adopt sandwich rules~\citep{NN_ICCV2019_Yu} to train the largest, smallest, and randomly sampled SubNets together.
(4) We normalize the reconstructed matrix $\rmW'(\vs,\vr)$ with a scaling factor $\sqrt{Var(\rmW)/Var(\rmW')}$ to match the target variance of $\rmW$ and avoid statistical instability.

\noindent\textbf{Per-Tensor Rank Selection via Evolutionary Search: $\vr^*$}.~
We randomly sample 2,560 SubNets%
\begin{wrapfigure}[14]{rH}{0.7\columnwidth}
\begin{minipage}{0.7\textwidth}
    \centering
    \vspace{-10pt}
    \includegraphics[width=\columnwidth]{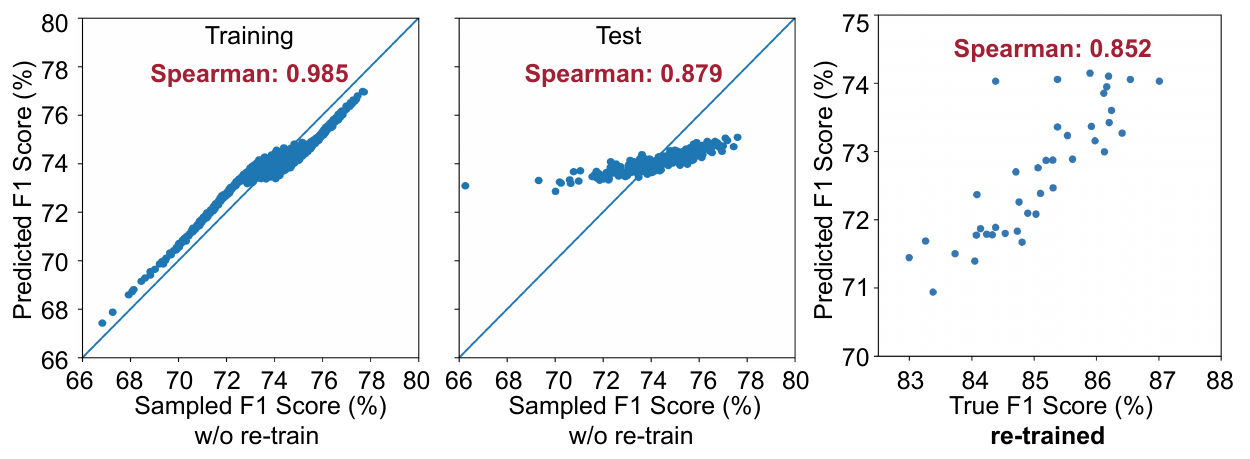}
    \vspace{-15pt}
    \caption{\textit{Left, Middle}: Our accuracy predictor shows high prediction fidelity and generalization on BERT-base SQuADv-1.1.
    \textit{Right}: Predicted F1 is an accurate proxy of the ground-truth re-trained F1.}
    \label{fig:AccuracyPredictor}
\end{minipage}
\end{wrapfigure}
from the Rank SuperNet and train a random forest $\calP: (\vs,\vr)\rightarrow \texttt{Acc}$ to predict the validation accuracy based on the factorization settings, which is a fast proxy to reduce validation cost.
In Fig.~\ref{fig:AccuracyPredictor}, we observe a high fidelity (high Spearman correlation) between the predicted and the re-trained F1.
In addition to accuracy maximization, network hardware cost must also be considered.
The total energy $E_{tot}$ and runtime $T_{tot}$ of the model is simply the sum of all layer costs.
We use evolutionary search to find the optimized factorization settings: $\min_{\vr}~(1-\texttt{Acc})(E_{tot}\cdot T_{tot})^{\gamma}$, where $\gamma$ is empirically set to 0.25.

\begin{wrapfigure}[13]{rH}{0.42\columnwidth}
\begin{minipage}{0.42\textwidth}
    \centering
    \vspace{-10pt}
    \includegraphics[width=\columnwidth]{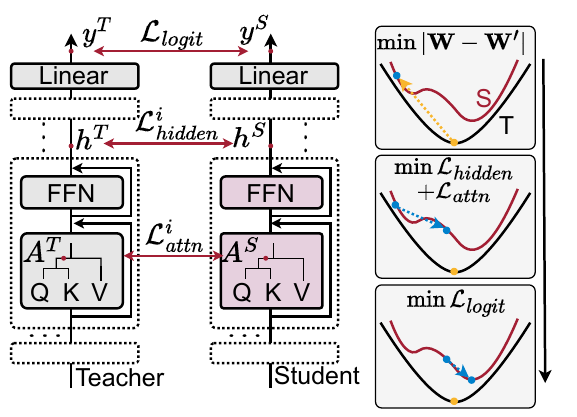}
    \caption{Layer-wise and logit distillation for factorized Transformers.}
    \label{fig:Distillation}
\end{minipage}
\end{wrapfigure}

\subsubsection{\texttt{Level} 3: Trainability Boost with Two-stage Distillation}
\label{sec:Distillation}
Since decomposition errors will accumulate through layers, re-training is necessary to recover accuracy.
However, optimization of factorized tensors is difficult, making trainability a bottleneck for factorized Transformers.
To solve this issue, we propose a two-stage distillation flow.
First, we perform optimal layer-wise projection to find the tensor decomposition that minimizes matrix approximation error, $\min_{\rmW_i'}\sum_i\|\rmW_i-\rmW'_i\|_{\calF}$.
Then we distill the layer-wise knowledge from the teacher $T$ to the factorized student $S$ both on the attention maps $A$ and hidden states $h$ on each Transformer block.
\begin{equation}
    \small
    \label{eq:LayerDistill}
    \begin{aligned}
    &\calL_{attn}+\calL_{hidden}=\sum_i \calL_{attn}^i+\calL_{hidden}^i\\
    &=\sum_i\texttt{CosEmbed}(A^S_i,A^T_i)+\texttt{CosEmbed}(h^S_i,h^T_i)
    \end{aligned}
\end{equation}
After the layer-wise alignment, we only apply last-layer logit distillation to provide more optimization freedom for the student,
\begin{equation}
    \centering
    \label{eq:LogitDistill}
    \calL_{logit}=\frac{1}{2} \mathcal{L}_{KL}(y^S/\tau,y^T/\tau)+\frac{1}{2}\calL_{CE}(y^S,y^T).
\end{equation}

\section{Results}
\label{sec:ExperimentalResults}
\subsection{Experiment setup}
\label{sec:ExpSetup}
\textbf{Datasets, Models, and Training Settings}.~ 
We search the decomposition settings on BERT-base/DistilBERT with the question-and-answer dataset SQuAD-v1.1 and evaluate on SQuAD and SST-2 datasets.
We use the original model fine-tuned on SQuAD as the teacher model.
We searched 3 variants, from \name-a1 to \name-a3, with different energy-delay product (EDP) to cover different design points.
Breakdown on the compression ratio, runtime, and energy cost of \name variants are in Appendix~\ref{fig:AppendixBreakdown} and \ref{fig:AppendixLatencyEnergyBreakdown}.
We evaluate three representative tensor decomposition methods: TTM, Tucker, and CP.
We follow the standard BERT fine-tuning settings.
Please see Appendix~\ref{sec:AppendixSuperNetTrain}, \ref{sec:AppendixRankSearch}, and \ref{sec:AppendixKDSettings} for detailed settings of the SuperNet training, evolutionary search, and knowledge distillation.

\textbf{Hardware Settings.}
We use Timeloop~\citep{NN_ISPASS2019_Parashar} as the mapper and hardware cost simulator, with energy models based on 5nm technology.
Detailed architecture configurations of our customized \texttt{Simba-L} accelerator and simulation settings can be found in Appendix~\ref{sec:AppendixCostSimulation}.

\subsection{Main Results}
\label{sec:MainResults}
\noindent\textbf{Results on BERT-base SQuAD.v1-1}.~
We compare our searched factorization settings with (1) the original fine-tuned BERT, (2) \texttt{SR-Manual}: manually-selected shape and rank settings, and (3) \texttt{S(Ours)-R(TensorLy)}: searched optimal tensorization shapes and heuristic ranks by TensorLy~\citep{tensorly} based on the target compression ratio.
In Table~\ref{tab:CompareResults} and Fig.~\ref{fig:MainResults}, \name achieves the best performance-efficiency Pareto front, surpassing manual and heuristic tensor decomposition baselines.

With TTM decomposition, \name improves the F1 score by +0.65\% with 8\% fewer parameters and 8.8\% lower hardware cost on average.
The compact \name-a1 benefits the most from our heterogeneous per-tensor rank settings and significantly outperforms manual and heuristic decomposition with \textbf{+1.6\%} higher F1 scores.
However, compared to the original BERT, we note that TTM-factorized BERT is not very hardware-efficient, as the TTM optimal contraction path reconstructs the weight matrix and performs the standard linear operation.

\begin{table}[]
\resizebox{\columnwidth}{!}{
\begin{tabular}{c|ccc|ccc|ccc}
\toprule
                        & \multicolumn{3}{c|}{TTM}                                                                                                         & \multicolumn{3}{c|}{Tucker}                                                                                                      & \multicolumn{3}{c}{CP}                                                                                                           \\ \cline{2-10} 
\multirow{-2}{*}{Model} & \begin{tabular}[c]{@{}c@{}}\#Params \\ (M)$\downarrow$\end{tabular} & F1 (\%)$\uparrow$        & \begin{tabular}[c]{@{}c@{}}EDP \\ ($\mu J\cdot s$)$\downarrow$\end{tabular} & \begin{tabular}[c]{@{}c@{}}\#Params \\ (M)$\downarrow$\end{tabular} & F1 (\%)$\uparrow$        & \begin{tabular}[c]{@{}c@{}}EDP \\ ($\mu J\cdot s$)$\downarrow$\end{tabular} & \begin{tabular}[c]{@{}c@{}}\#Params \\ (M)$\downarrow$\end{tabular} & F1 (\%)$\uparrow$        & \begin{tabular}[c]{@{}c@{}}EDP \\ ($\mu J\cdot s$)$\downarrow$\end{tabular} \\ \midrule
BERT-base               & 109.50                                                  & 88.16          & 34.31                                                 & 109.50                                                  & 88.16          & 34.31                                                 & 109.50                                                  & 88.16          & 34.31                                                 \\
\midrule
SR-Manual-1             & 35.04                                                   & 83.60          & 24.27                                                 & 38.10                                                   & 81.90          & 5.63                                                  & 43.13                                                   & 68.95          & 4.48                                                  \\
SR-Manual-2             & 38.71                                                   & 84.74          & 25.86                                                 & 39.71                                                   & 82.31          & 6.85                                                  & 54.27                                                   & 79.22          & 7.57                                                  \\
S(Ours)-R(TensorLy)-1   & 36.15                                                   & 83.89          & 24.42                                                 & 35.81                                                   & 80.43          & 4.06                                                  & 46.38                                                   & 85.81          & 5.45                                                  \\
S(Ours)-R(TensorLy)-2   & 39.78                                                   & 85.49          & 27.18                                                 & 39.48                                                   & 82.00          & 4.93                                                  & 55.52                                                   & 87.21          & 9.37                                                  \\
\rowcolor[HTML]{EFEFEF} 
\textbf{\name-a1}        & \textbf{41.60}                                          & \textbf{86.01} & \textbf{25.06}                                        & \textbf{42.14}                                          & \textbf{83.41} & \textbf{4.91}                                         & \textbf{48.88}                                          & \textbf{87.11} & \textbf{5.99}                                         \\ \midrule
SR-Manual-3             & 42.37                                                   & 86.05          & 28.87                                                 & 42.64                                                   & 83.45          & 7.87                                                  & 60.88                                                   & 81.96          & 10.70                                                 \\
S(Ours)-R(TensorLy)-3   & 43.33                                                   & 85.84          & 29.05                                                 & 42.63                                                   & 82.96          & 5.66                                                  & 60.31                                                   & 87.74          & 9.62                                                  \\
\rowcolor[HTML]{EFEFEF} 
\textbf{\name-a2}        & \textbf{46.42}                                          & \textbf{86.71} & \textbf{28.93}                                        & \textbf{50.96}                                          & \textbf{85.51} & \textbf{7.45}                                         & \textbf{59.08}                                          & \textbf{87.79} & \textbf{9.01}                                         \\ \midrule
SR-Manual-4             & 50.90                                                   & 86.99          & 32.57                                                 & 52.94                                                   & 85.05          & 12.21                                                 & 69.12                                                   & 84.31          & 12.40                                                 \\
SR-Manual-5             & 60.68                                                   & 87.29          & 39.25                                                 & 59.17                                                   & 86.50          & 17.00                                                 & 83.98                                                   & 86.59          & 18.50                                                 \\
S(Ours)-R(TensorLy)-4   & 51.76                                                   & 86.64          & 33.55                                                 & 51.87                                                   & 85.19          & 9.27                                                  & 68.79                                                   & 87.62          & 14.40                                                 \\
S(Ours)-R(TensorLy)-5   & 61.45                                                   & 87.74          & 49.27                                                 & 62.26                                                   & 86.76          & 11.90                                                 & 82.43                                                   & 88.14          & 18.00                                                 \\
\rowcolor[HTML]{EFEFEF} 
\textbf{\name-a3}        & \textbf{54.68}                                          & \textbf{87.36} & \textbf{32.72}                                        & \textbf{63.62}                                          & \textbf{86.35} & \textbf{10.40}                                        & \textbf{74.15}                                          & \textbf{87.91} & \textbf{14.50}                                        \\ \midrule
\textbf{Avg. Improv.}   & \textbf{-8.15\%}                                        & \textbf{+0.65} & \textbf{-8.80\%}                                      & \textbf{+0.09\%}                                        & \textbf{+1.03} & \textbf{-30.28\%}                                     & \textbf{-9.07\%}                                        & \textbf{+3.45} & \textbf{-25.30\%}\\\bottomrule
\end{tabular}
}
\caption{Compare our \name-series with baseline decomposition methods on BERT-base SQuAD-v1.1 in terms of \#Params, F1 score, and energy-delay product (EDP) across three decomposition methods.}
\label{tab:CompareResults}
\vspace{-5pt}
\end{table}

With Tucker decomposition, \name-series boosts the F1 score by \textbf{+1.03} with \textbf{+30\%} higher efficiency than handcrafted settings that typically reshape the matrix to a high-order tensor (e.g., order 6 or 8)~\citep{NN_CVPR2021_Yin,NN_IJCNN2018_Tjandra}.
However, this widely-used heuristic turns out to be much less efficient than the lower-order tensorization found by \name, because the high-order Einsum equation includes many 2-operand Einsums with tiny tensor dimensions ($\ll$32), resulting in low hardware utilization.
Though the accuracy per parameter of Tucker is slightly lower than TTM, its accuracy-to-EDP ratio is 3-5$\times$ higher than TTM. 

CP decomposition is usually disfavored due to unstable optimization~\citep{silva2008cp}, which is also evidenced by manual CP decomposition.
In contrast, our search framework finds an efficient CP tensorization shape and is able to recover accuracy through re-training.
\name-series overall can boost the F1 score by \textbf{+3.45\%} with \textbf{25.3\%} less hardware cost.
Compared to the original BERT, our searched \name-a1 can maintain accuracy ($\leq$1\% drop) with \textbf{5.7$\times$} higher hardware efficiency.

\begin{wrapfigure}[8]{rH}{0.5\columnwidth}
\begin{minipage}{0.5\textwidth}
\centering
\vspace{-10pt}
\resizebox{\columnwidth}{!}{
\begin{tabular}{l|cc|cc|cc}
\toprule
\multicolumn{1}{c|}{\multirow{2}{*}{Model}} & \multicolumn{2}{c|}{TTM} & \multicolumn{2}{c|}{Tucker} & \multicolumn{2}{c}{CP} \\
\multicolumn{1}{c|}{}                       & F1 (\%)$\uparrow$     & EDP$\downarrow$    & F1(\%)      & EDP$\downarrow$      & F1 (\%)$\uparrow$    & EDP$\downarrow$   \\ \midrule
BERT-base                                   & 91.74      & 5.21        & 91.74       & 5.21          & 91.74     & 5.21       \\ \midrule
\name-a1                                     & 90.02      & 4.24        & 87.27       & 0.45          & 91.40     & 0.50       \\
\name-a2                                     & 90.90      & 5.65        & 88.42       & 0.69          & 91.51     & 0.79       \\
\name-a3                                     & 91.20      & 7.34        & 89.91       & 1.02          & 91.17     & 1.34       \\ \bottomrule
\end{tabular}
}
\vspace{-5pt}
\captionof{table}{Evaluation of \name on SST-2 with decomposition settings searched on SQuAD-v1.1.}
\label{tab:CompareDatasets}
\end{minipage}
\end{wrapfigure}

\noindent\textbf{Results on BERT-base SST-2}.~
We re-train \name-variants on SST-2 to evaluate the generalization of the decomposition settings searched on SQuAD.
In Table~\ref{tab:CompareDatasets}, we observe that when adapted to a new downstream task with a smaller sequence length (128), our searched Tucker and CP factorization can still largely maintain the F1 score with \textbf{4-10$\times$} higher hardware efficiency.

\noindent\textbf{Results on DistilBERT SQuADv-1.1 and SST-2}.~
Our tensor decomposition method can be applied to compact Transformers as an orthogonal compression technique.
Based on DistilBERT~\citep{NN_NeurIPS2019_Sanh}, a 6-layer compact version of BERT-base, we searched three \name-variants in Table~\ref{tab:CompareDistilBert}.
Our \name-series can achieve comparable F1 scores with \textbf{5.7$\times$} higher efficiency on SQuAD-v1.1.
On SST-2, \name-series can maintain the accuracy while saving \textbf{3-10$\times$} hardware cost.

\begin{table}[htp]
    \centering
\resizebox{0.95\textwidth}{!}{
\begin{tabular}{c|cccccc|cccccc}
\toprule
\multicolumn{1}{l|}{\multirow{3}{*}{}} & \multicolumn{6}{c|}{SQuADv1.1}                                                 & \multicolumn{6}{c}{SST-2}                                                     \\
\multicolumn{1}{l|}{}                  & \multicolumn{2}{c}{TTM} & \multicolumn{2}{c}{Tucker} & \multicolumn{2}{c|}{CP} & \multicolumn{2}{c}{TTM} & \multicolumn{2}{c}{Tucker} & \multicolumn{2}{c}{CP} \\ \cmidrule{2-13} 
\multicolumn{1}{l|}{}                  & F1$\uparrow$          & EDP$\downarrow$       & F1$\uparrow$           & EDP$\downarrow$         & F1$\uparrow$          & EDP$\downarrow$       & Acc$\uparrow$         & EDP$\downarrow$       & Acc$\uparrow$          & EDP$\downarrow$         & Acc$\uparrow$        & EDP$\downarrow$       \\ \midrule
DistilBERT                             & 86.90       & 8.58      & 86.90        & 8.58        & 86.90       & 8.58      & 91.97       & 1.30      & 91.97        & 1.30        & 91.97      & 1.30      \\ \midrule
\name-a1                                & 85.26       & 6.71      & 83.66        & 1.72        & 87.09       & 1.50      &  91.40           & 1.23      &  90.30            & 0.17        &   91.28         & 0.13      \\
\name-a2                                & 86.42       & 7.76      & 85.21        & 2.56        & 87.89       & 2.26      &  90.60           & 1.64      &  91.30            & 0.23        &  91.06          & 0.20      \\
\name-a3                                & 86.92       & 8.74      & 86.15        & 3.21        & 88.18       & 3.63      &   90.60          & 2.10      &  90.60            & 0.29        &  91.63          & 0.34      \\ \bottomrule
\end{tabular}
}
\vspace{-5pt}
\caption{Compare three \name-variants with DistilBERT-base on SQuAD-v1.1 and SST-2 datasets.}
\label{tab:CompareDistilBert}
\vspace{-10pt}
\end{table}

\subsection{Discussions}
\label{sec:Discussion}
\noindent\textbf{Ablation on SuperNet Training Techniques}.~
We perform ablation studies on the Rank SuperNet%
\begin{wrapfigure}[10]{rH}{0.27\columnwidth}
\begin{minipage}{0.27\textwidth}
    \centering
    \vspace{-12pt}
    \includegraphics[width=\columnwidth]{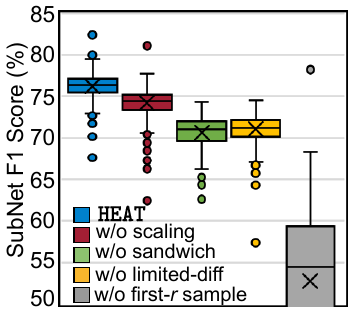}
    \vspace{-16pt}
    \caption{Ablation on SuperNet training.}
    \label{fig:SupernetTrainAblation}
\end{minipage}
\end{wrapfigure}

training techniques in Fig.~\ref{fig:SupernetTrainAblation}.
We individually remove one technique at a time and show the F1 score distribution of 1,024 SubNets.
With all training techniques, our \name framework achieves the best SuperNet convergence with the highest SubNet F1 scores.

\begin{figure}[htp]
    \centering
    \vspace{-10pt}
    \includegraphics[width=\columnwidth]{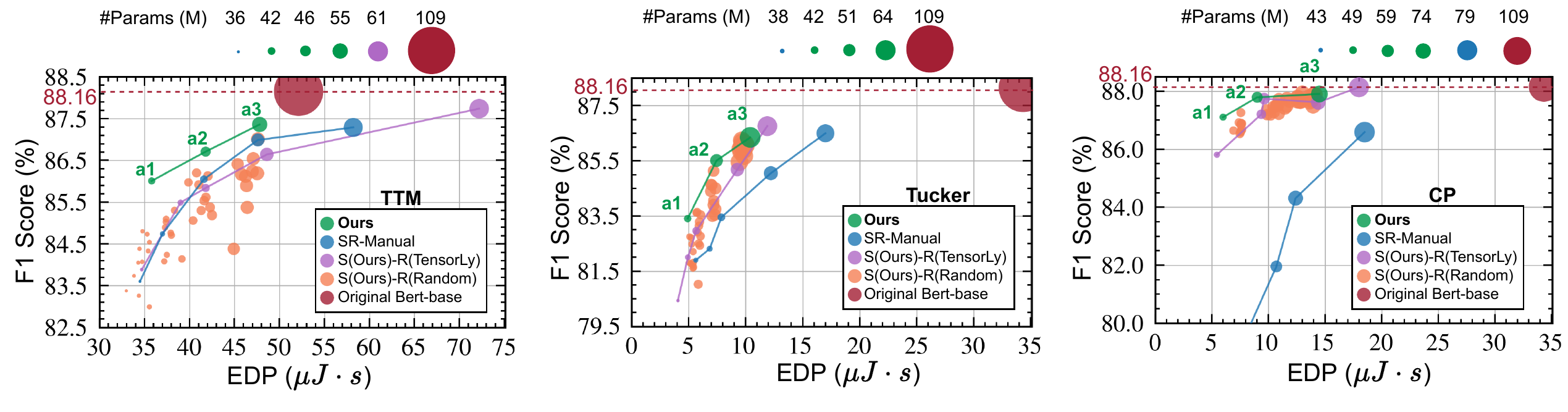}
    \vspace{-15pt}
    \caption{Comparison of our searched factorization with baselines in the accuracy vs. EDP space on BERT-base SQuAD-v1.1.}
    \label{fig:MainResults}
\end{figure}

\noindent\textbf{Evolution vs. Random Rank Settings}.~
In Fig.~\ref{fig:MainResults}, we further fine-tune 40 SubNets on SQuAD-v1.1 with our searched tensorization shape and randomly sampled per-tensor ranks, denoted as \texttt{S(Ours)-R(Random)}.
We can observe that (1) even with random ranks, our searched shape can still outperform manually selected shapes;
(2) and the evolutionary search effectively explores the Pareto front of the rank distribution. 

\noindent\textbf{Ablation on Re-Training Recipes}.~
We compare different re-training methods in Table~\ref{tab:CompareRetrain} and visualize four representative attention maps, which are key indicators of the model representability.
Direct re-training with cross-entropy loss suffers from severe performance loss due to optimization difficulty, and the attention map can barely be recovered.
One-stage distillation is not effective since the layer-wise loss is not compatible with logit distillation loss, especially in the later optimization stage.
When we decouple layer-wise and logit distillation, especially when attention and hidden state distillation are combined in the first stage, we observe significant improvement in accuracy.
The abundant patterns in the visualized attention maps are mostly recovered.
\begin{table}[htp]
\resizebox{\columnwidth}{!}{
\begin{tabular}{l|c|c|cccc|ccc}
\toprule
\multirow{2}{*}{}   & \multirow{2}{*}{BERT-base}                                                         & w/o KD  & \multicolumn{4}{c|}{One-Stage KD}                                   & \multicolumn{3}{c}{Two-Stage KD}                                                                          \\
                     &                                                        & Finetune (16) & L (16) & A+L (16) & H+L (16) & A+H+L (16) & A(8) $\rightarrow$ L(8) & H(8) $\rightarrow$ L(8) & \textbf{A+H(8) $\rightarrow$ L(8)} \\ \midrule
\multicolumn{1}{c|}{\begin{tabular}[c]{@{}c@{}}Attention\\ Map\end{tabular}} & \begin{minipage}{.125\textwidth}
      \includegraphics[width=1.0\linewidth, height=19mm]{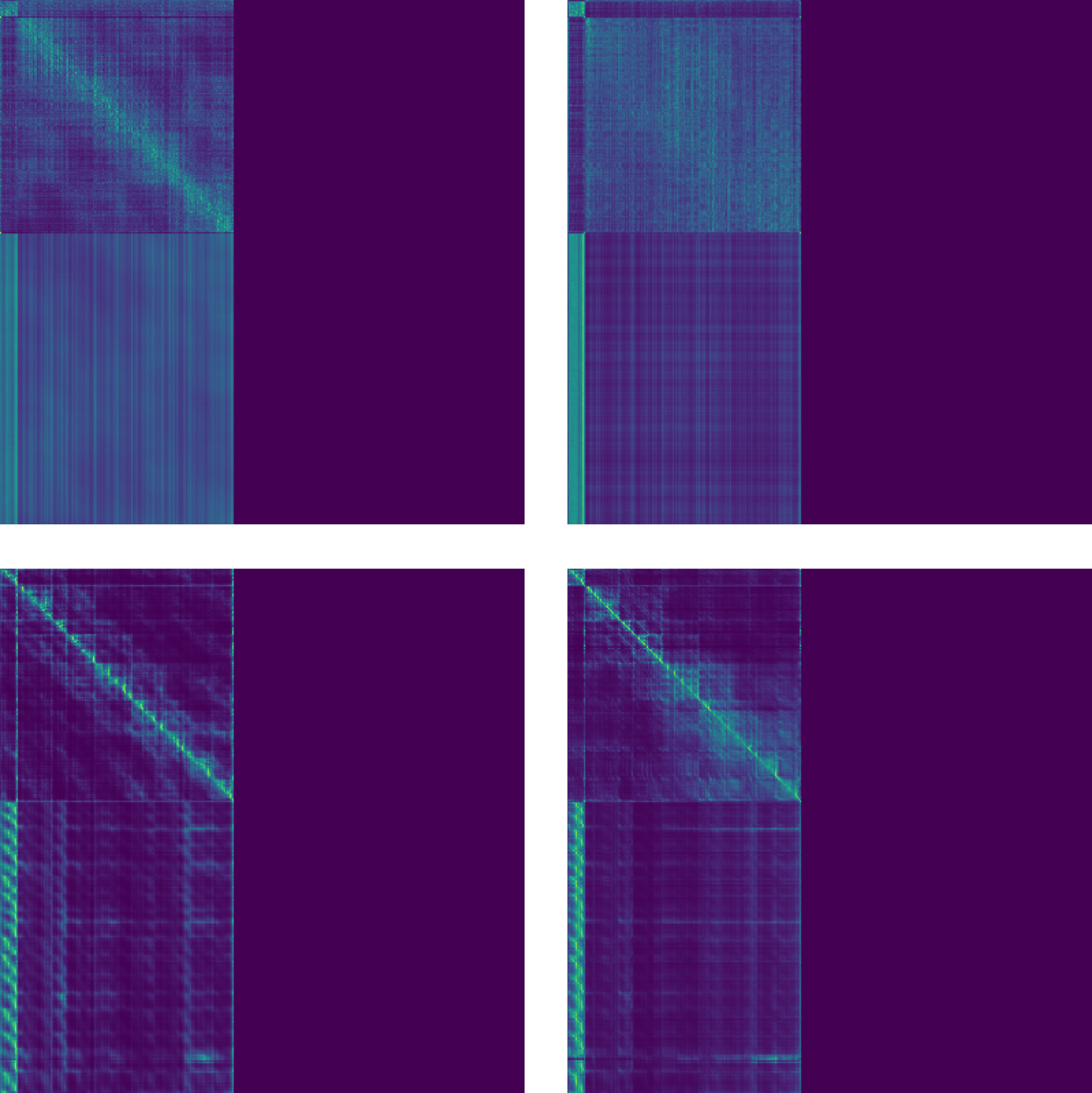}
    \end{minipage} & \begin{minipage}{.125\textwidth}
      \includegraphics[width=1.0\linewidth, height=19mm]{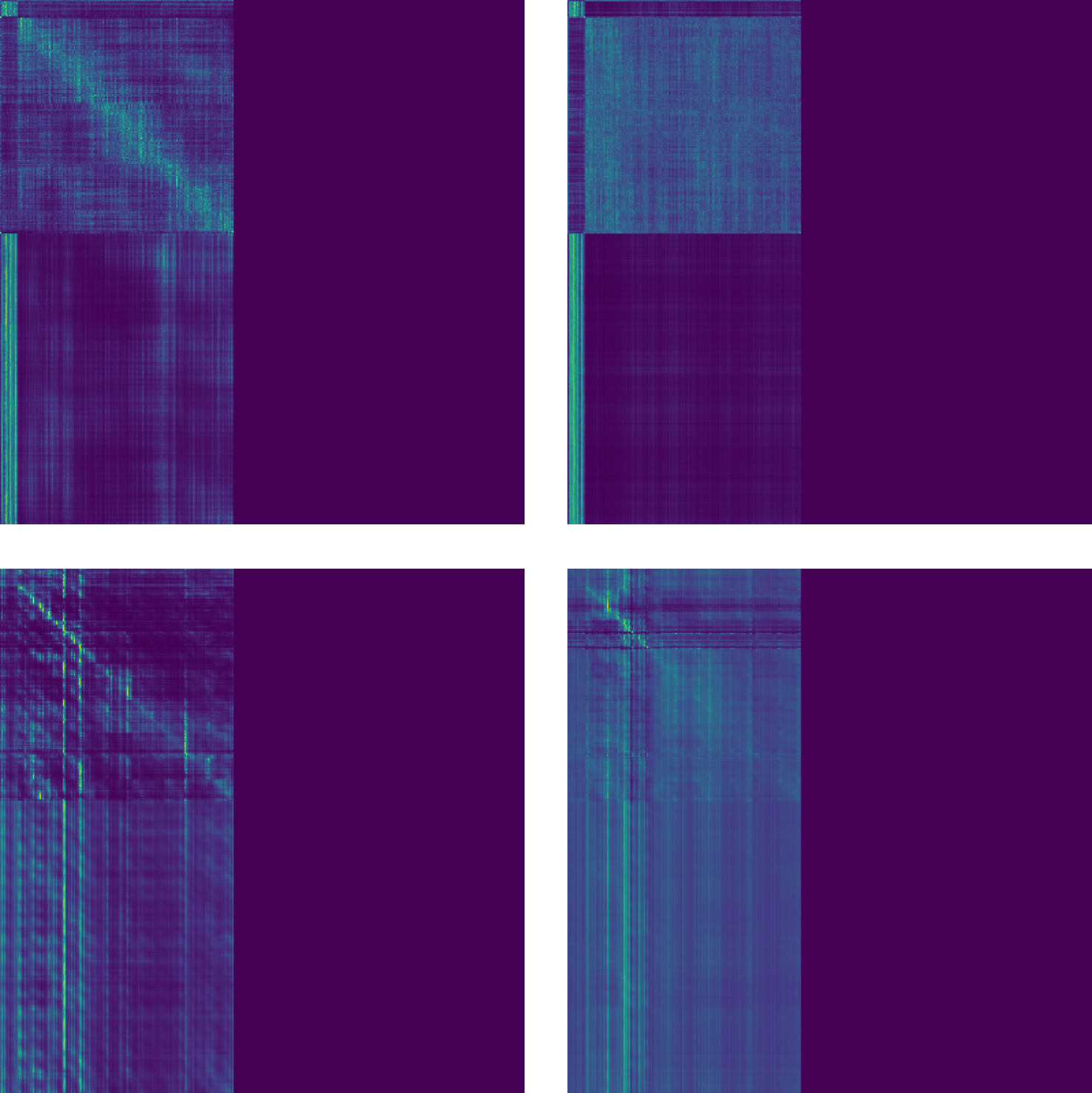}
    \end{minipage}        &  \begin{minipage}{.125\textwidth}
      \includegraphics[width=1.0\linewidth, height=19mm]{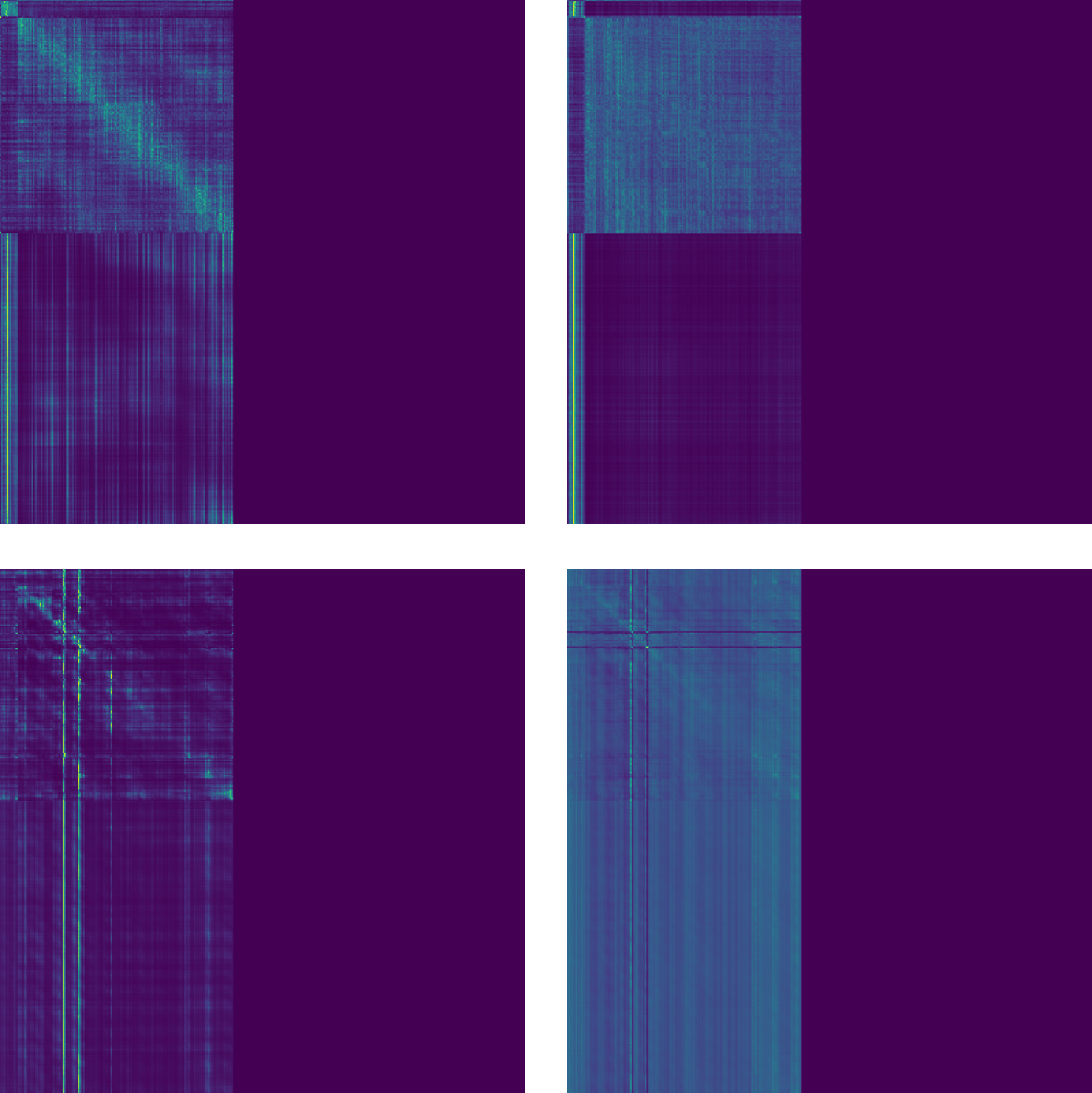}
    \end{minipage}          &  \begin{minipage}{.125\textwidth}
      \includegraphics[width=1.0\linewidth, height=19mm]{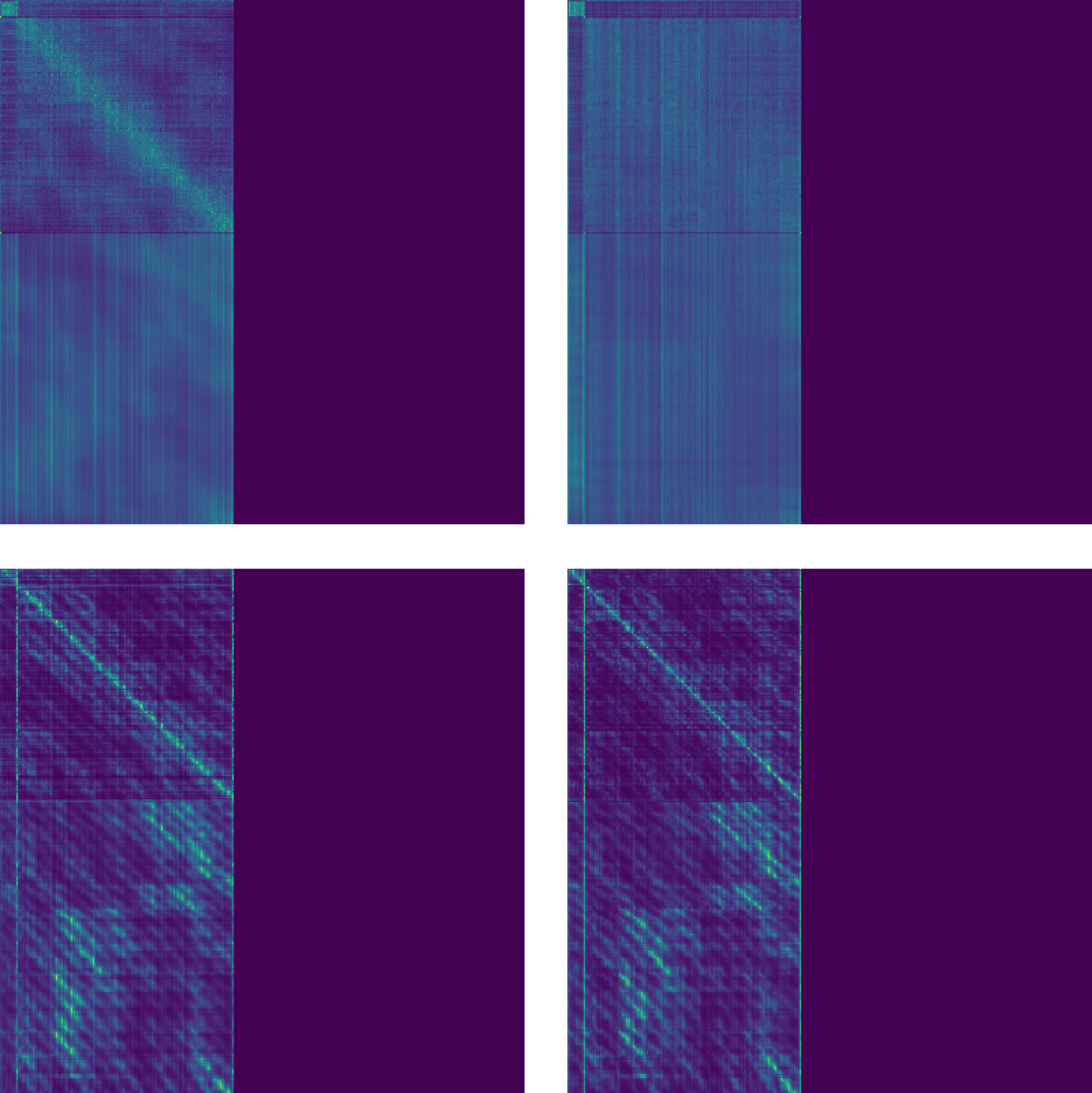}
    \end{minipage}               &   \begin{minipage}{.125\textwidth}
      \includegraphics[width=1.0\linewidth, height=19mm]{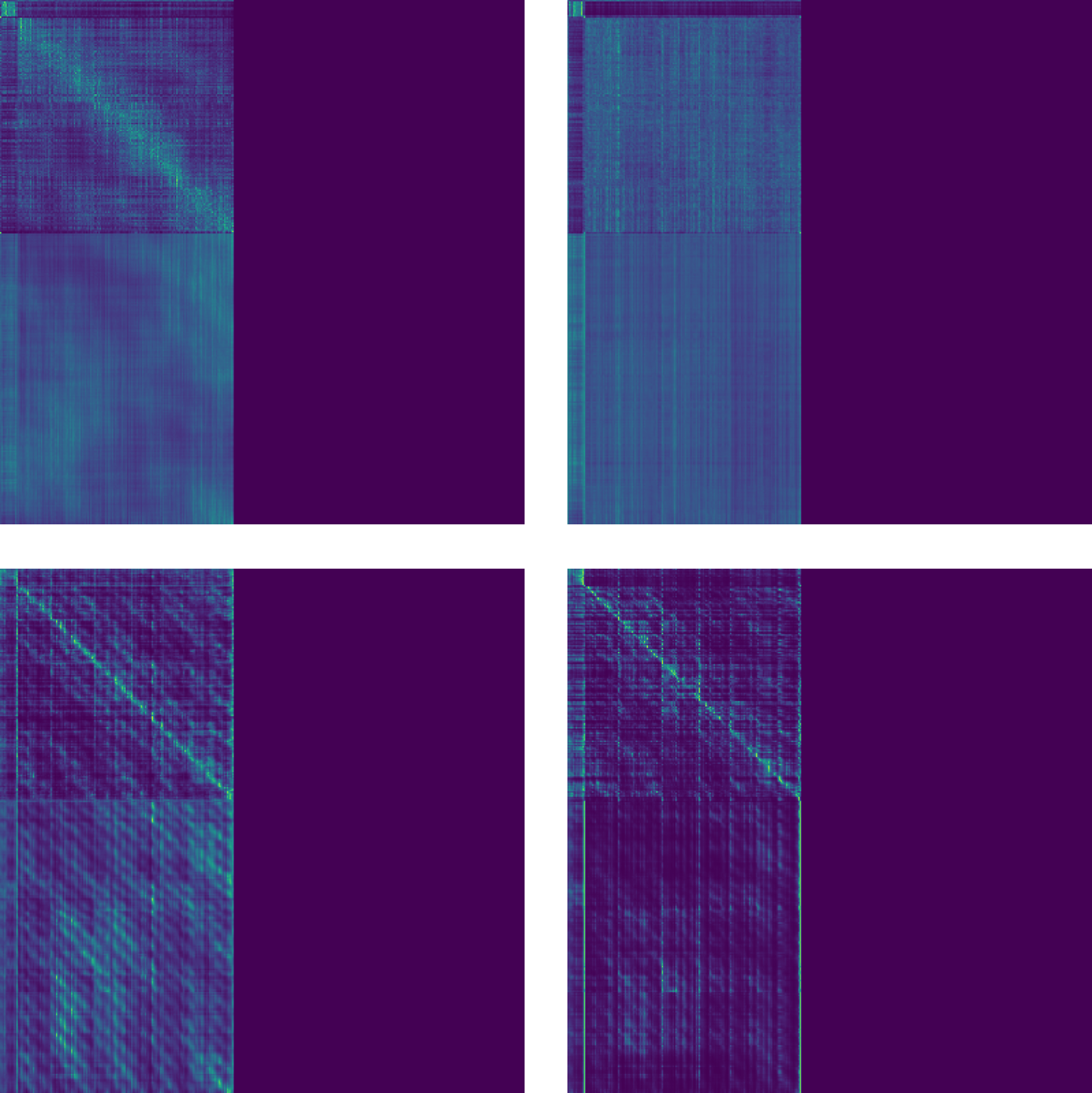}
    \end{minipage}             &   \begin{minipage}{.125\textwidth}
      \includegraphics[width=1.0\linewidth, height=19mm]{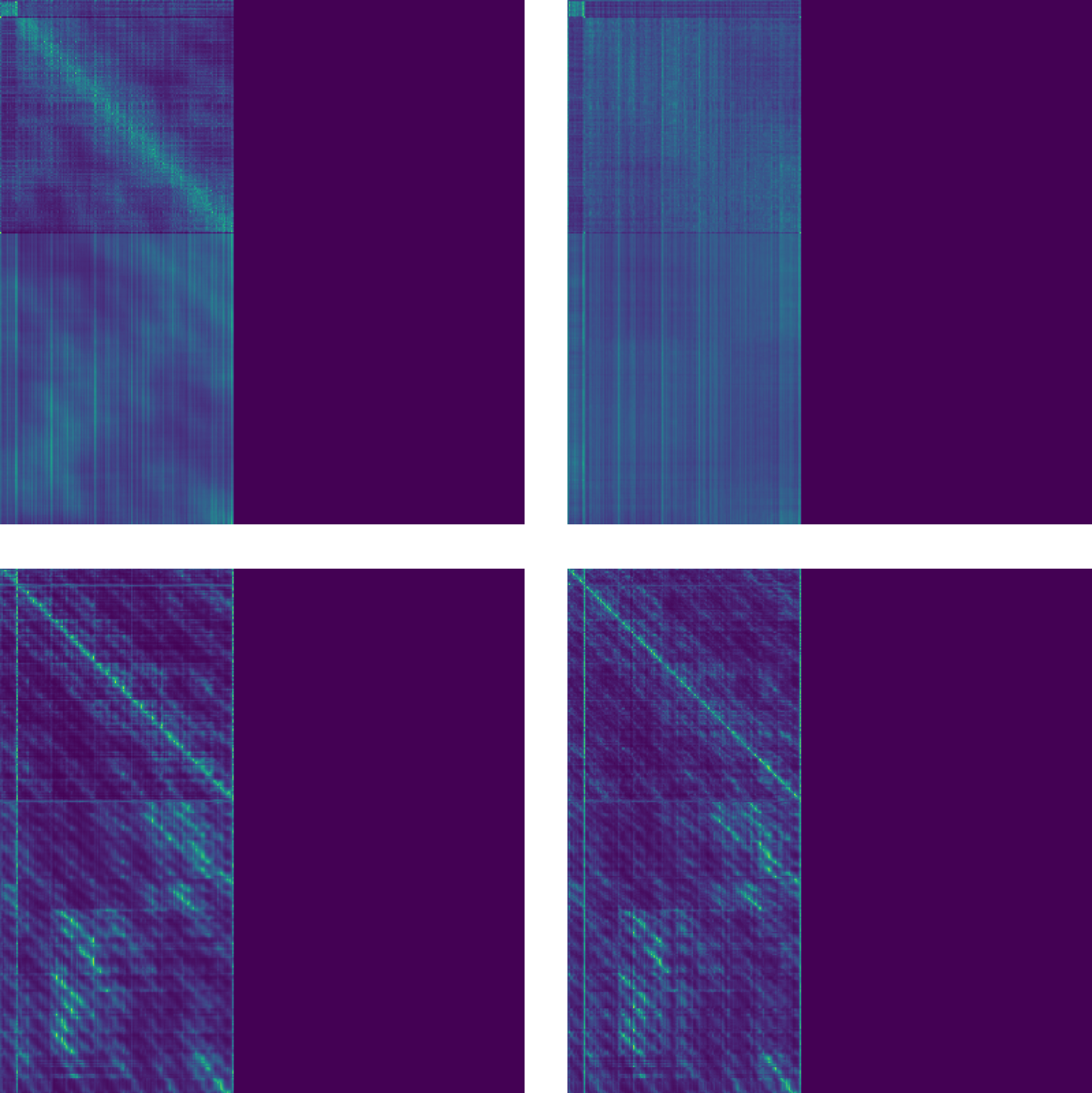}
    \end{minipage}                  &   \begin{minipage}{.125\textwidth}
      \includegraphics[width=1.0\linewidth, height=19mm]{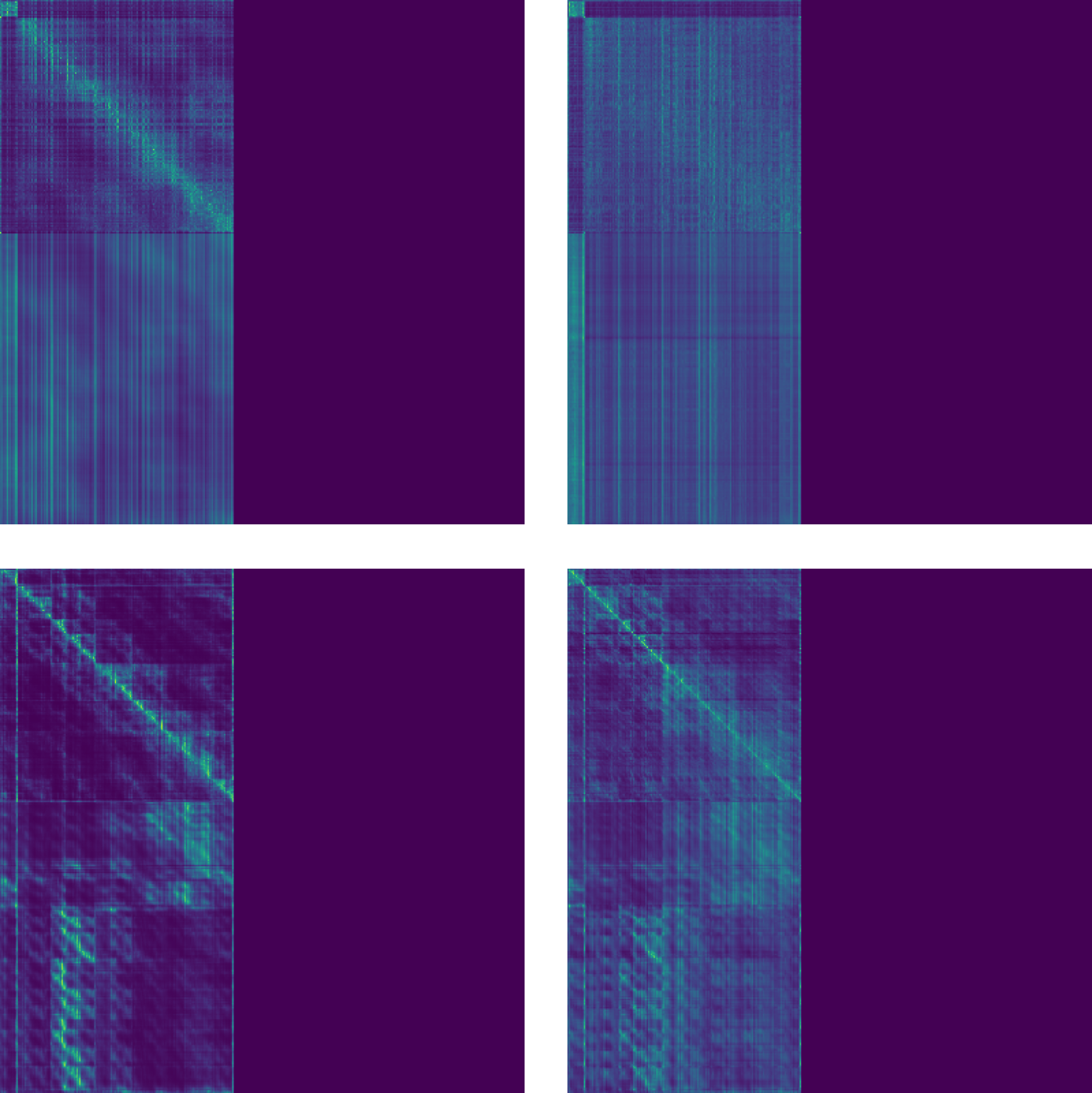}
    \end{minipage}                               &   \begin{minipage}{.125\textwidth}
      \includegraphics[width=1.0\linewidth, height=19mm]{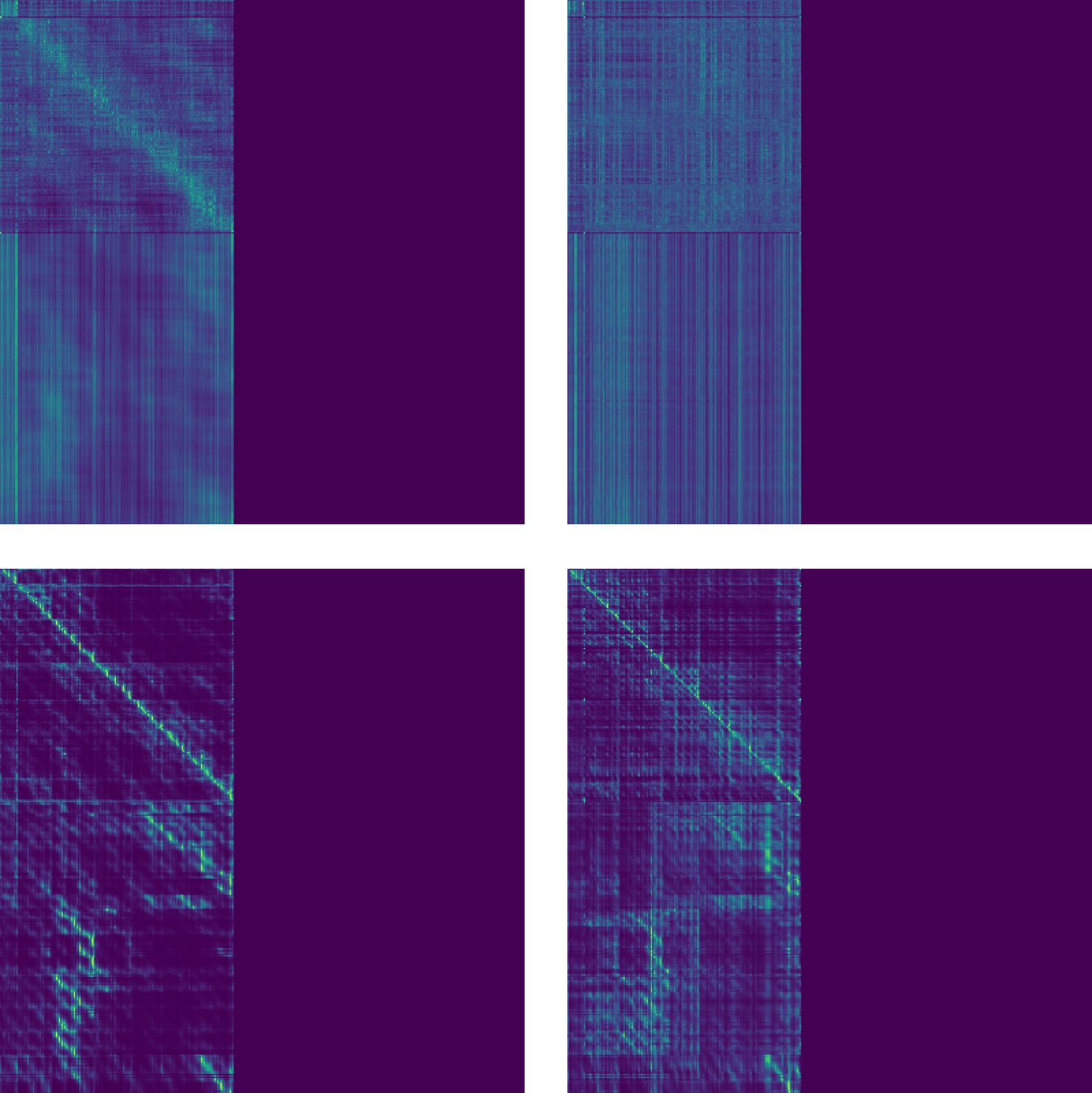}
    \end{minipage}                              & \begin{minipage}{.125\textwidth}
      \includegraphics[width=1.0\linewidth, height=19mm]{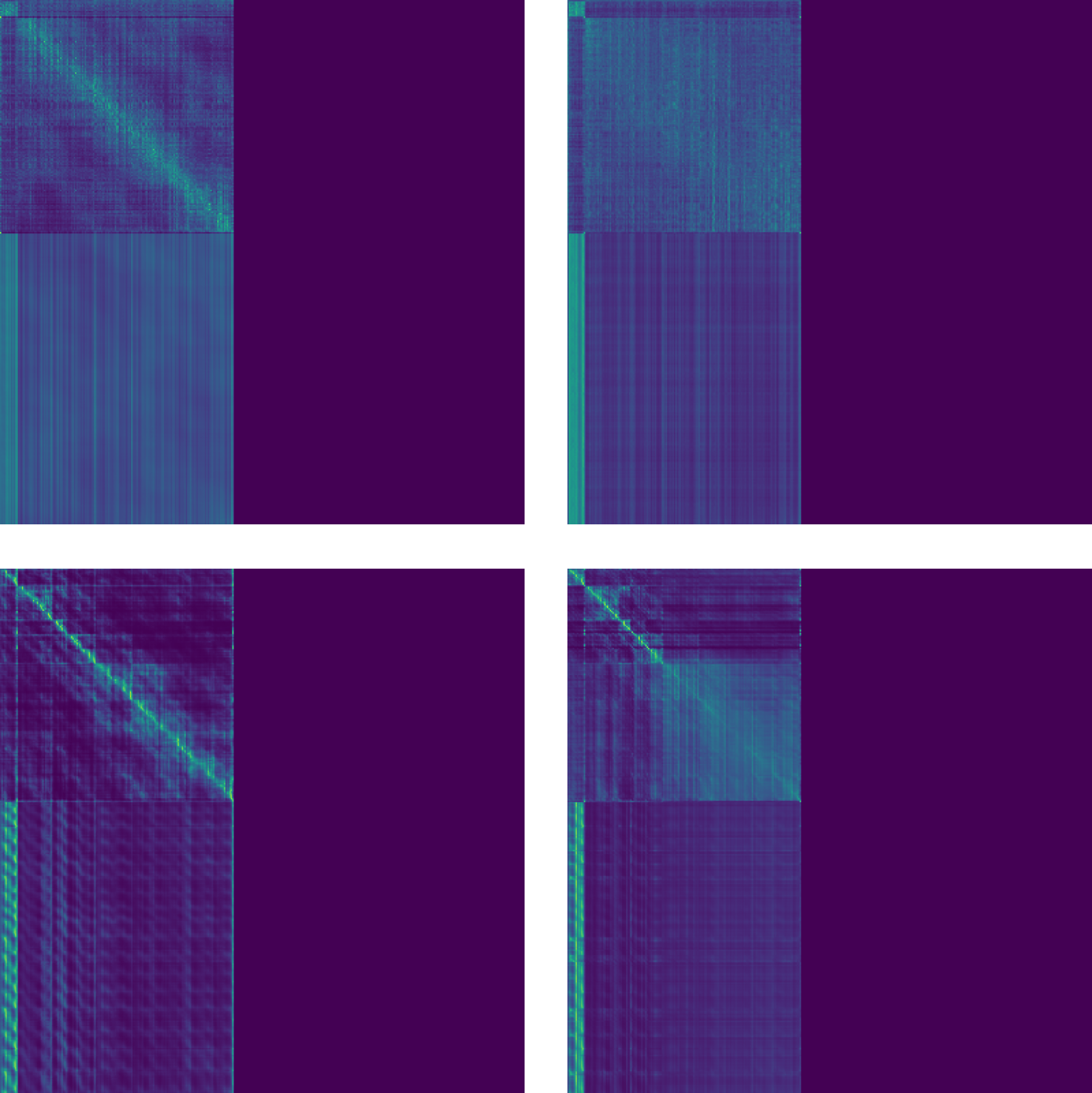}
    \end{minipage}                                     \\ \midrule
\multicolumn{1}{c|}{F1 (\%)}    & 88.16                                             &  79.11       &  79.22          &  78.32               &  75.89              & 81.15                    &  85.99                                &  73.22                               &  \textbf{87.36}                                    \\ \bottomrule
\end{tabular}
}
\caption{Compare different re-training recipes of TTM-factorized \name-a3 on SQuAD-v1.1.
All methods are trained for 16 epochs in total.}
\label{tab:CompareRetrain}
\end{table}

\section{Related Work}
\label{sec:Background}
\noindent\textbf{Compression with Tensor Decomposition}.~
In the literature on Transformer compression, tensor decomposition is applied to compress the embedding layers~\citep{NN_EMNLP2020_Hrinchunk,NN_NeruIPS2018_Zhen,NN_MLSys2021_Yin}.
However, we \textit{do not decompose the embedding matrix} due to limited hardware efficiency benefits (See Appendix~\ref{sec:AppendixEmbedding}).
Multi-linear attention~\citep{NN_NeurIPS2019_Ma} was proposed based on block-term tensor decomposition to achieve parameter-efficient Transformer.
TIE~\citep{NN_ISCA2019_Deng} applied TT format to all linear matrices and proposed a new hardware design to minimize redundant computations.
Recently, hybrid compression approaches have achieved strong results in Transformer compression.
Low-rank decomposition was applied with weight pruning~\citep{NN_AAAI2021_Zhen} or quantization to slim down BERT by 7.5$\times$.
Our work delves deeply into tensor decomposition, and the proposed \name framework can be jointly applied with other orthogonal methods, which is one promising future direction.

\noindent\textbf{Optimization of Factorized NNs}.~
Knowledge distillation was used to transfer the expressivity from a pre-trained teacher Transformer to the compact student model~\citep{NN_AAAI2021_Zhen, NN_NeurIPS2019_Sanh}.
Bayesian tensorized NNs are put forward to automatically determine the rank in low-rank decomposition without manual settings~\citep{NN_SIAM2020_Hawkins}.
Our search framework is more scalable than the Bayesian method and considers the real accuracy and hardware cost instead of just matrix decomposition error and compression ratios.

\section{Conclusion}
\label{sec:Conclusion}
In this work, we explore the large design space of hardware-efficient tensor decomposition and present \name, an automatic decomposition framework for Transformer model compression.
We move beyond conventional manual factorization focusing only on compression ratios or computations.
We consider hardware cost in the optimization loop and efficiently find expressive and hardware-efficient tensorization shapes.
Our SuperNet-based one-shot rank search flow can efficiently generate optimized per-tensor decomposition rank settings.
We employ a two-stage distillation flow to solve the trainability bottleneck of factorized Transformers and significantly boost their task performance.
Experiments show that \name reduces up to 5.7$\times$ energy-delay product on our customized accelerator with less than 1.1\% accuracy drop.
Compared to manual and heuristic tensor decomposition methods, our searched \name-variants show 1-3\% higher accuracy with $\sim$30\% less hardware cost on average.

\newpage

% {\small
% \bibliographystyle{iclr2023_conference}
% \bibliography{./ref/Top_sim,./ref/NN,./ref/NP,./ref/ALG, ./ref/addition,./ref/tensor}
% }

\newpage
\appendix
\renewcommand{\thepage}{A\arabic{page}}  
\renewcommand{\thesection}{A\arabic{section}}   
\renewcommand{\thetable}{A\arabic{table}}   
\renewcommand{\thefigure}{A\arabic{figure}}

\section{Memory Access Types for Fused Einsum}
\label{sec:AppendixFusedEinsum}
To implement fused einsum, we separate all nodes in the tensor contraction path $p$ into 4 categories according to their memory access types, as shown in Fig.~\ref{fig:FusedEinsum}.
\begin{itemize}[leftmargin=*]
\setlength{\itemindent}{0.5em}
    \item \texttt{Type 1}: Load dynamic (D) input from DRAM and static (S) weights from the global buffer (GB), and write the intermediate tensor back to GB for data reuse.
\end{itemize}

\begin{itemize}[leftmargin=*]
\setlength{\itemindent}{0.5em}
    \item \texttt{Type 2}: Only read/write from/to GB without costly DRAM transaction.
    \item \texttt{Type 3}: Load both operands from GB and write the final results back to DRAM.
    \item \texttt{Type 4}: Load one dynamic input from DRAM and write the final results to DRAM.
\end{itemize}
Different node types are simulated with corresponding memory access constraints in Timeloop, so we can implement a fused einsum without unnecessary DRAM access.

\section{Hardware Cost Simulation Settings}
\label{sec:AppendixCostSimulation}
To construct the hardware cost table $\calT$, we construct the tensorization shape candidate spaces in Table~\ref{tab:AppendixShapeSpace}.
For each shape, we collect all candidate ranks that are multiples of 32 or 8 and satisfy the compression ratio constraints.
We use \texttt{Timeloop} with TSMC 5 nm energy model to simulate the fused einsum operation corresponding to each shape-rank pair.
Our \texttt{Simba-L} architecture has a 2.91 MB global buffer and 32 PEs.
Each PE has 32 32-KB weight buffers, one 64-KB input buffer, 32 384-B accumulation buffers, and 1024 8-bit MAC units.
The hardware mapping objective of \texttt{Timeloop-mapper} is to minimize the energy-delay product.
Based on the simulated hardware cost table $\calT$, we use the \href{https://github.com/tommyod/paretoset}{\texttt{paretoset}} library to automatically select the Pareto optimal tensorization shape for the subsequent rank search flow.
\begin{table}[htp]
\caption{Tensorization shape search space for the query, value, key, and projection weight matrices with three factorization methods.}
\resizebox{\columnwidth}{!}{
\begin{tabular}{c|c|c|c|cc}
\hline
Model  & Candidate orders & \begin{tabular}[c]{@{}c@{}}Top-k \\ Candidate shapes\end{tabular} & Candidate ranks & \begin{tabular}[c]{@{}c@{}}Compression Ratio\\ Limits\end{tabular} & \#Candidates \\ \hline
TTM    & 6, 8, 10          & 3                                                                 & Multiple of 32  & 0.35$\sim$0.5                                                       & 3235         \\
Tucker & 4, 6, 8          & 3                                                                 & Multiple of 8   & 0.35$\sim$0.5                                                       & 4754         \\
CP     & 2, 3             & 3                                                                 & Multiple of 32  & 0.35$\sim$0.5                                                       & 44           \\ \hline
\end{tabular}
}
\label{tab:AppendixShapeSpace}
\end{table}

\section{Rank SuperNet Training Settings}
\label{sec:AppendixSuperNetTrain}
With the searched optimal tensorization shape $\vs^*$, we construct a Rank SuperNet with maximum ranks following a $\sim$60\% target compression ratio.
We use the original fine-tuned BERT-base as the teacher model and launch the 10-epoch logit distillation flow to train the Rank SuperNet.
We use Adam optimizer with a learning rate of 3e-5 and a linear decay schedule.
In the limited difference technique, we restrict the maximum allowed rank change across iterations to 3.
We use a sandwich rule with one largest SubNet, one smallest SubNet, and two randomly sampled SubNets.

\section{Per-tensor Rank Search Settings}
\label{sec:AppendixRankSearch}
With the trained Rank SuperNet, we uniformly sample 2560 SubNets with the largest and smallest SubNets and evaluate their validation F1 scores on a 5\% validation set.
We use 95\% SubNet evaluation data to train a random forest ensemble model as the accuracy predictor.
The model is an AdaBoostRegressor with 100 ExtraTreeRegressor, each tree regressor containing 60 decision trees with a maximum depth of 10.

In the evolutionary search stage, we use 200 populations with 40 parents, 80 mutations with 50\% mutation probability, and 80 crossovers.
After 100 steps, we obtain the optimal per-tensor rank settings.

\section{Knowledge Distillation Settings}
\label{sec:AppendixKDSettings}
We use a two-stage knowledge distillation flow to train the model with the searched ($\vs^*$, $\vr^*$) settings.
We use the original BERT-base as the teacher model.
We first launch an 8-epoch layer-wise distillation flow with attention and hidden state mapping with a constant learning rate of 3e-5 for TTM and Tucker, 1e-5 for CP.
Then, we launch an 8-epoch logit distillation flow with an initial learning rate of 1e-5 on SQuAD-v1.1 and 6e-6 on SST-2 and a linear decay rate with 10\% warm-up.

\section{Breakdown of Searched \name Variants}
\label{sec:AppendixBreakdown}
\begin{figure}
    \centering
    \includegraphics[width=\columnwidth]{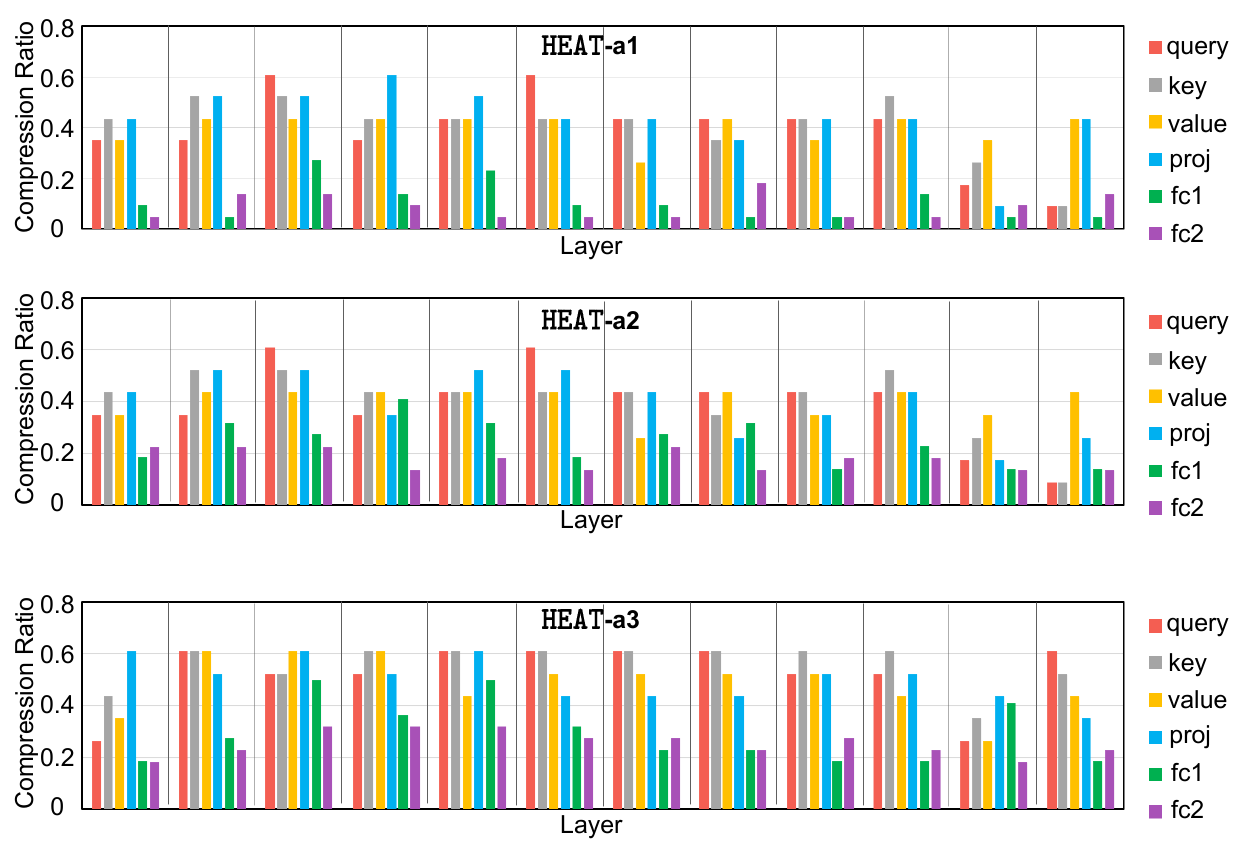}
    \caption{Compression ratio breakdown on each matrix in our \name-variants with TTM decomposition.}
    \label{fig:AppendixBreakdown}
\end{figure}

\noindent\textbf{Compression Ratio}.~
We plot the compression ratio breakdown of our searched \name-variants in Fig.~\ref{fig:AppendixBreakdown}.
We can observe that deeper layers tend to have higher redundancy and thus have fewer parameters.
Feedforward networks tend to have a lower compression ratio (fewer parameters) than query/value/key matrices.

\begin{figure}
    \centering
    \includegraphics[width=\columnwidth]{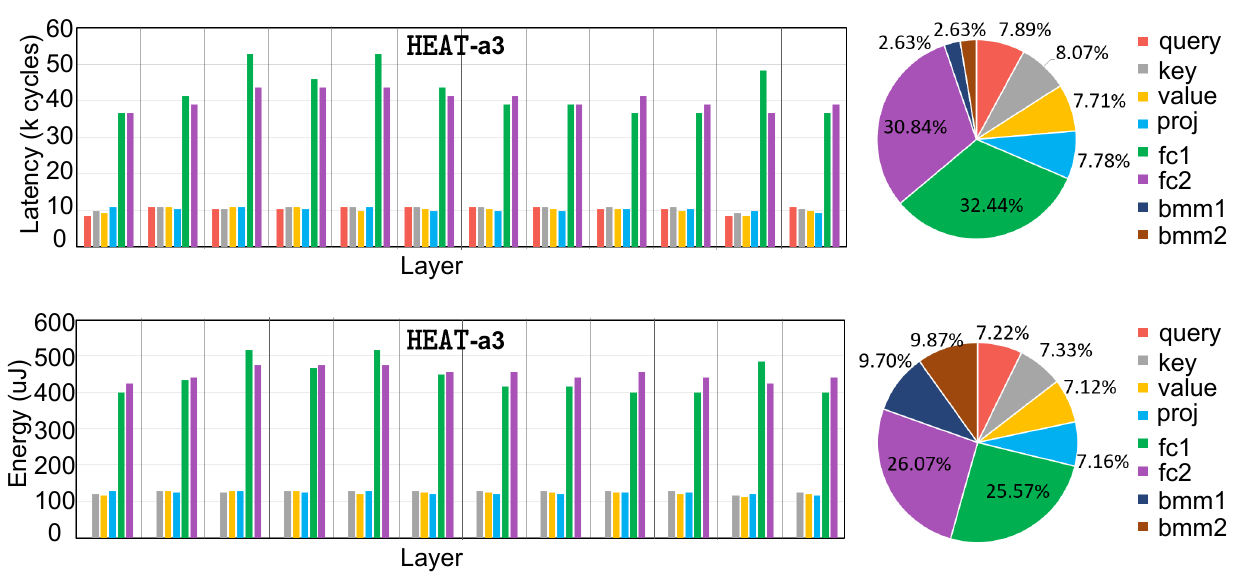}
    \caption{Latency (\textit{Top}) and energy (\textit{Bottom}) breakdown of \name-a3.}
    \label{fig:AppendixLatencyEnergyBreakdown}
\end{figure}

\noindent\textbf{Latency and Energy}.~
In Fig.~\ref{fig:AppendixLatencyEnergyBreakdown}, the fully-connected (FC) layers in FFNs have lower compression ratios but nearly 4$\times$ higher latency than other linear layers.
The batched matrix multiplication (BMM) in attention operations, i.e., $\mathbf{Q}\mathbf{K}^T$ and $\mathbf{AV}$, only take around 5.3\% total latency and 19.5\% total energy in the entire network, which validates that the most costly operations in Transformer are indeed linear layers.

\section{Decomposition on Embedding Layers}
\label{sec:AppendixEmbedding}
Some prior work applies low-rank decomposition on the embedding layer in Transformer models and claims it can save parameters.
However, when off-chip DRAM capacity is not a limiting factor, this embedding layer decomposition comes with a non-trivial accuracy drop and no hardware efficiency benefits.
Indexing the original look-up table only contains DRAM read without extra computations.
In contrast, indexing factorized tensors is very costly and requires tensor dot-product among all decomposed core tensors.
The computation overhead far outweighs the saved storage capacity.
Therefore, we do not decompose the embedding layers.

\end{document}